\documentclass[11pt]{article}
\usepackage[a4paper,margin=1in]{geometry}
\RequirePackage{tgtermes}
\RequirePackage{newtxtext}
\RequirePackage{newtxmath}
\RequirePackage{bm}
\RequirePackage{endnotes}

\usepackage{algorithm}
\usepackage{algpseudocode}
\usepackage{marginnote}
\usepackage{tikz}
\usepackage{multirow}
\usepackage{multicol}
\usepackage{booktabs}
\usepackage{amsmath,amsfonts}
\usepackage{subcaption}
\usepackage{multicol}
\usepackage{placeins}
\usepackage{amsmath}
\usepackage{tcolorbox}
\usepackage{wrapfig}
\usepackage{xcolor}    
\usepackage{url}
\usepackage{enumitem}

\usepackage{pdfpages}
\definecolor{myblue}{RGB}{35, 76, 143} 
\usepackage{hyperref}
\hypersetup{
    colorlinks=true,
    linkcolor=myblue, 
    citecolor=myblue, 
    filecolor=magenta,     
    urlcolor=black          
}

\usepackage{endnotes}

\let\footnote=\endnote

\newcommand{\FigureNoteStylere}[1]{\footnotesize \linespread{1.0}\selectfont \parbox{\textwidth}{\textit{Note. } #1}}
\newtheorem{remark}{Remark}
\newtheorem{definition}{Definition}

\newtheorem{proposition}{Proposition}
\usepackage{natbib}
\usepackage{ulem}
\usepackage{diagbox}
\usepackage{listings}
\newcommand{\subrefparen}[1]{(\subref{#1})}
\lstdefinestyle{mystyle}{
    keywordstyle=\color{blue!70},			
    commentstyle=\color{red!50!green!50!blue!50},		
    numberstyle=\tiny\color{gray},		
    stringstyle=\color{purple},
    basicstyle=\ttfamily\tiny, 
    breakatwhitespace=true,         
    breaklines=true,	                
    captionpos=b,                    
    keepspaces=true,                 
    showspaces=false,
    numbers=left,
    showstringspaces=false,		
    showtabs=false,                  
    tabsize=2,
    frame=single, 
    aboveskip=0pt, 
    belowskip=0pt 
}

 \bibpunct[, ]{(}{)}{,}{a}{}{,}%

\title{Learning Virtual Machine Scheduling in Cloud Computing through Language Agents}
\author{
\parbox{0.95\textwidth}{\centering
Jiehao Wu\textsuperscript{a,\#}, Ziwei Wang\textsuperscript{b,\#}, Junjie Sheng\textsuperscript{a}, Wenhao Li\textsuperscript{c}, Xiangfeng Wang\textsuperscript{d,*}, Jun Luo\textsuperscript{b,*} \\
\small \textsuperscript{a}\,School of Computer Science and Technology, East China Normal University, Shanghai 200062, China \\
\small \textsuperscript{b}\,Antai College of Economics and Management, Shanghai Jiao Tong University, Shanghai 200030, China \\
\small \textsuperscript{c}\,School of Computer Science and Technology, Tongji University, Shanghai 200092, China \\
\small \textsuperscript{d}\,Key Laboratory of Mathematics and Engineering Applications, MoE, \\
\small East China Normal University, Shanghai 200062, China \\
\small \textsuperscript{\#}\,These authors contributed equally to this work. \\
\small \textsuperscript{*}\,Corresponding authors
}
}
\date{}

\begin{document}

\maketitle

\begin{abstract}
In cloud services, virtual machine (VM) scheduling is a typical Online Dynamic Multidimensional Bin Packing (ODMBP) problem, characterized by large-scale and nonstationary demands. 
Traditional optimization methods struggle to adapt to dynamic environments, existing learning-based methods often lack generalizability and interpretability, and domain-expert-designed heuristic approaches are constrained by rigid strategies.
To address these limitations, this paper proposes a hierarchical language agent framework named MiCo, which provides a large language model (LLM)-driven heuristic design paradigm for solving ODMBP. Specifically, ODMBP is formulated as a Semi-Markov Decision Process with Options (SMDP-Option), enabling dynamic scheduling through a micro-macro hierarchical architecture, i.e., Option Miner and Option Composer.
Option Miner utilizes LLMs to discover context-independent strategies through environment interactions. Option Composer uses LLMs to develop a context-aware composing strategy that integrates the context-independent strategies. 
Extensive experiments on a real-world dataset demonstrate that MiCo achieves a 96.9\% performance ratio under large-scale and nonstationary scenarios. It maintains high performance even under nonstationary request flows and different configurations.
\end{abstract}

\noindent\textbf{Funding:} The work of J. Luo was supported in part by the National Natural Science Foundation of China [72031006 and 72542012]. The work of W. Li was supported in part by the National Natural Science Foundation of China [62406270] and the STCSM Shanghai Rising-Star Program [24YF2748800].



%


\section{Introduction}\label{sec: introduction}
Cloud computing has emerged as the dominant model for delivering computing resources.
It allows users to access computing resources over the Internet, without managing the underlying infrastructure. This on-demand, scalable model has made cloud computing an attractive alternative for both business and individual consumers. According to \cite{gartner_web}, worldwide end-user spending on public cloud services was \$595.7 billion in 2024, and is expected to be \$723.4 billion in 2025. 
On the backend, service providers face the challenge of on-demand, dynamic resource allocation among virtual machines (VMs), commonly referred to as VM scheduling. 
As cloud infrastructure expands, the complexity and scale of VM scheduling increase, making it critical for operational efficiency and technological advancement.
Correspondingly, the operations management (OM) community has developed a strong interest in on-demand VM allocation because this issue is closely related to emergency patient allocation and customer management within the OM field \citep{chen2023cloud}. 
Efficiently managing VMs is crucial to maximizing physical machine (PM) utilization, reducing operational costs, and enhancing system performance.

\begin{figure}[htbp]
\centering
\caption{An Example of VM Scheduling.}\label{fig: sched illu}
\includegraphics[width=0.7\textwidth]{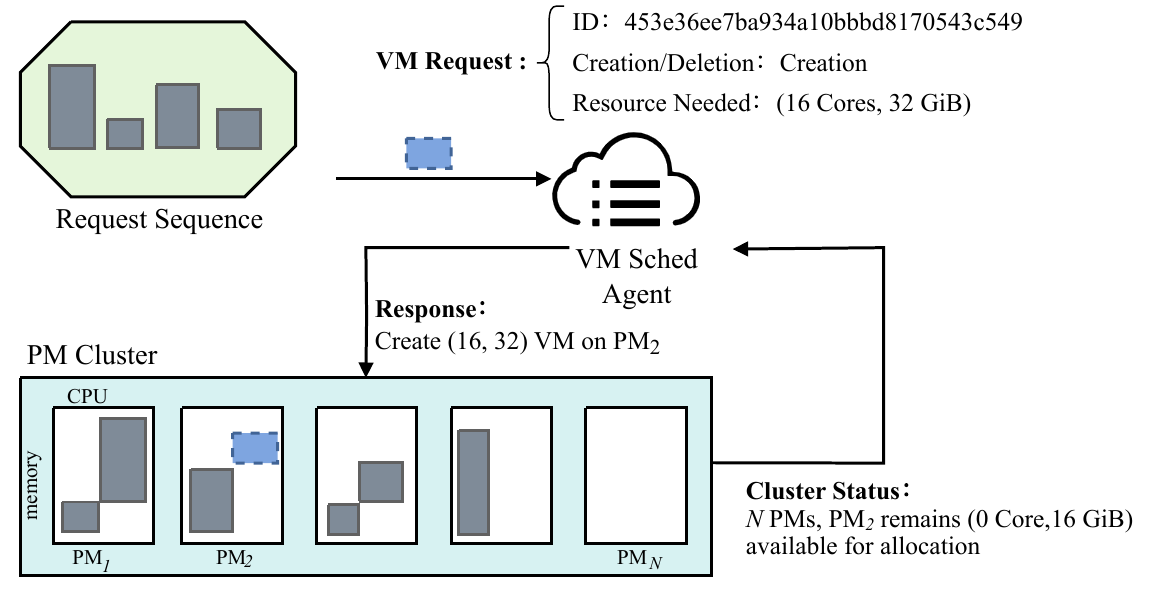}

\FigureNoteStylere{The process can be succinctly described as follows: The VM scheduling agent receives a sequence of VM requests, including creation or deletion commands and resource requirements. The agent then determines the appropriate PM for VM allocation. In this instance, a new VM is assigned to PM index 2. The decision is subsequently communicated to the PM cluster, which executes the resource allocation and VM creation accordingly.}
\end{figure}

\begin{figure}[htbp]
    \centering
    \caption{Characteristics of VM Requests over One Year Period.}
    \label{fig:VM-distribution}
    \vspace{2mm}
    \begin{subfigure}[b]{0.32\textwidth}
        \caption{\small }
        \label{fig:arrival-sub1}
        \includegraphics[width=\textwidth]{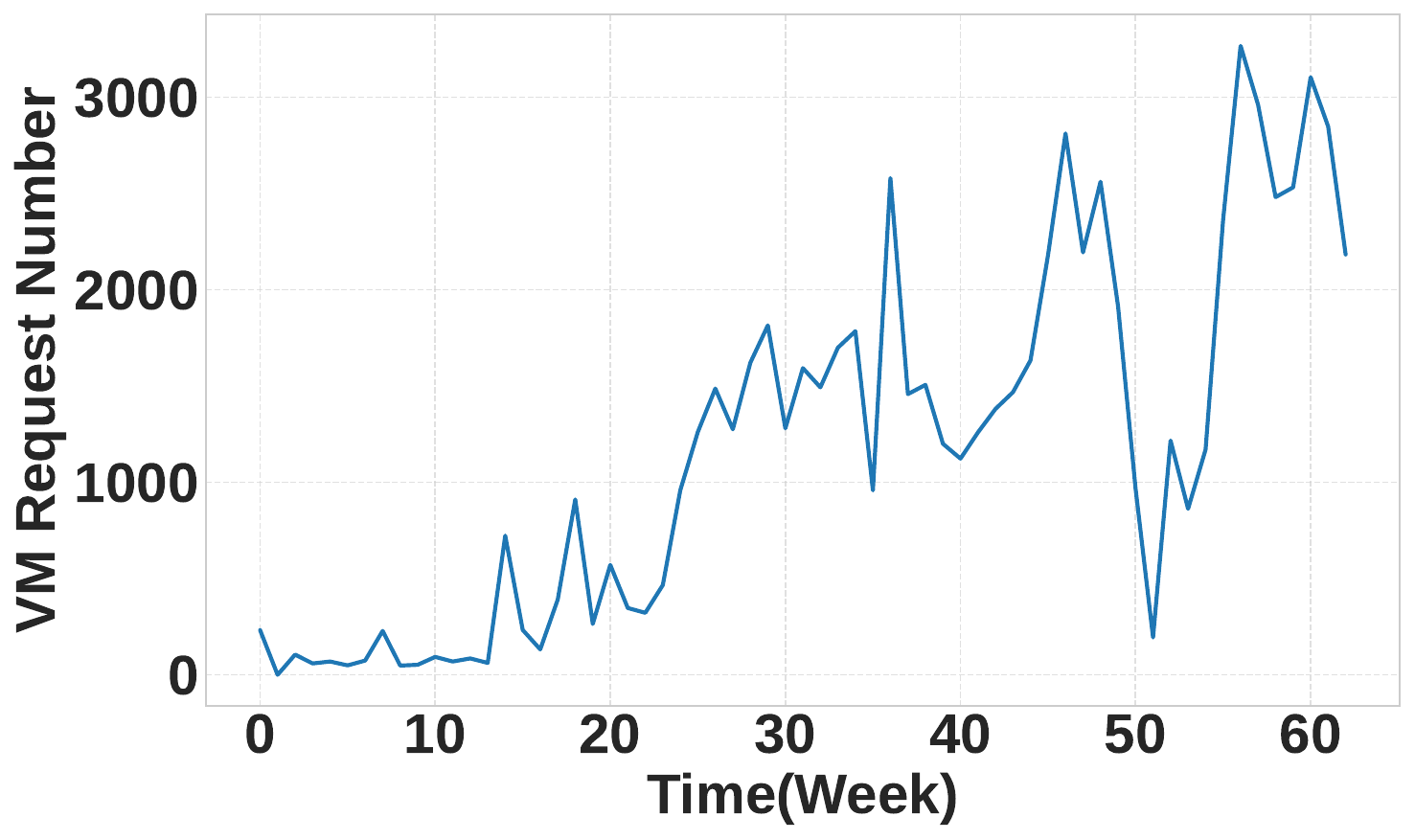}
    \end{subfigure}
    \begin{subfigure}[b]{0.32\textwidth}
        \caption{\small }
        \label{fig:duration-sub2}
        \includegraphics[width=\textwidth]{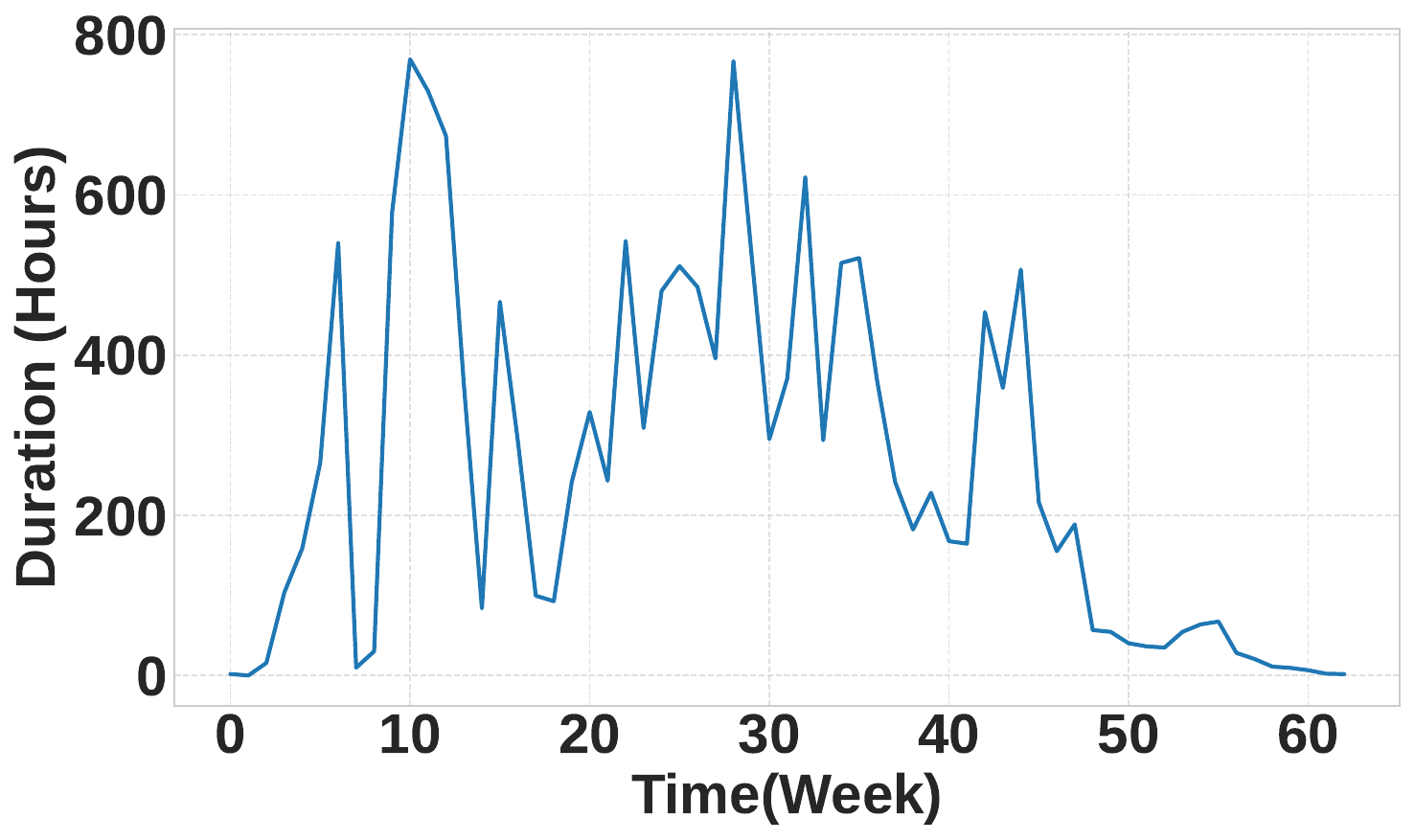}
    \end{subfigure}
    \begin{subfigure}[b]{0.32\textwidth}
        \caption{\small }
        \label{fig:size-sub3}
        \includegraphics[width=\textwidth]{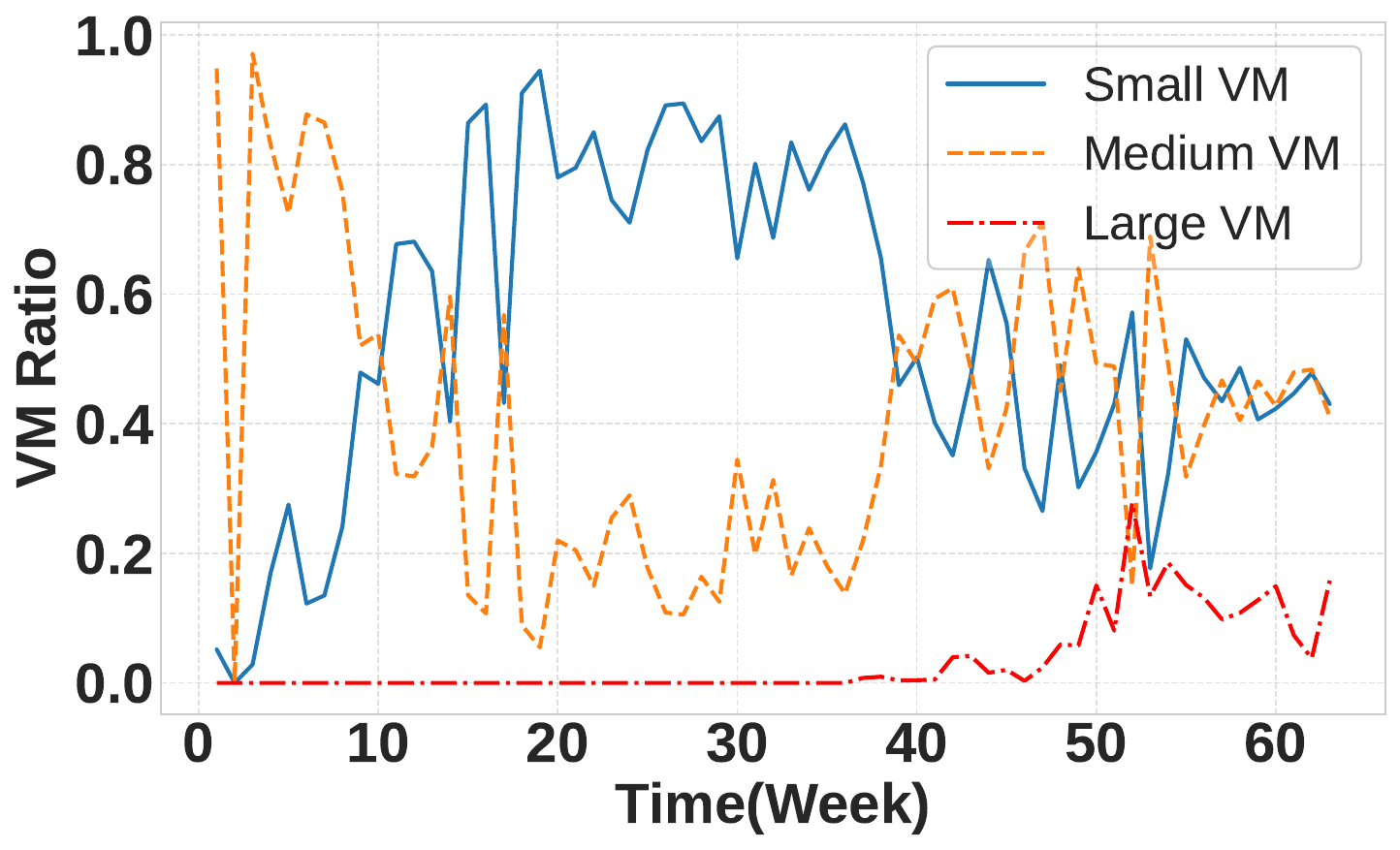}
    \end{subfigure}
    
    \FigureNoteStylere{(a) Number of VM Arrivals; (b) Average Duration Time; (c) Diverse Demand Size Distributions. We group VM requests by week, and calculate the total request arrival numbers per week, as shown in Figure \ref{fig:VM-distribution}\subrefparen{fig:arrival-sub1}. The average duration per week is displayed in Figure \ref{fig:VM-distribution}\subrefparen{fig:duration-sub2}. VMs are categorized into small, medium, and large based on CPU/memory dimensions, and the proportions of these request types over the total requests are depicted in Figure \ref{fig:VM-distribution}\subrefparen{fig:size-sub3}.}
\end{figure}

The VM scheduling problem can be framed as an \textbf{online dynamic multidimensional bin packing (ODMBP)} problem, a well-known NP-hard challenge. 
Figure \ref{fig: sched illu} illustrates the process of handling VM requests, aiming to optimally allocate each incoming VM to a suitable PM, thereby maximizing the total number of scheduled VMs. 
The problem is characterized by its online and dynamic nature: requests arrive sequentially with uncertain future information, and requests can arrive or leave at any time \citep{coffman1983dynamic}. 
Additionally, both VMs and PMs have multidimensional resource requirements such as CPU and memory \citep{christensen2017approximation}. 
Figure \ref{fig:VM-distribution} illustrates the \textbf{large-scale} and \textbf{nonstationary} features of the VM scheduling problem, as evidenced by statistical analysis based on a real-world VM trace dataset from Huawei Cloud, comprising approximately 125,000 VM requests over one year. 
The temporal fluctuations in VM resource requirements significantly expand the optimization search space, necessitating the development of an advanced algorithm to address the ODMBP problem under these conditions.

\subsection{Related Work}

Traditional approaches for solving the ODMBP problem typically fall into three categories: \textit{optimization-based}, \textit{learning-based}, and \textit{heuristic} algorithms. \textit{Optimization-based methods} can achieve exact optimal solutions in offline settings when item arrival sequences are fully known \citep{martello2000three, zhang2020branch, cote2021combinatorial}. In online settings, \cite{berg2017fast} and \cite{stolyar2013infinite} explored resource allocation with dynamic departures.
\textit{Learning-based approaches}, particularly reinforcement learning (RL), offer rapid response in online settings for large-scale ODMBP problems \citep{jiang2021learning, tang2024workflow}.
\textit{Heuristic algorithms} are widely adopted in industrial practice for their simplicity and efficiency. Classical methods like \textit{First-Fit}, \textit{Best-Fit}, and \textit{Next-Fit} provide foundational solutions with established worst-case bounds \citep{azar2013tight, azar2019tight}. Industry applications, such as Microsoft’s Protean, use rule-driven heuristics to optimize resource utilization and VM allocation \citep{hadary2020protean}. An extensive analysis of the literature associated with these methods is presented in Appendix \ref{secA4}.

However, despite their effectiveness in specific contexts, these traditional approaches face intrinsic limitations when applied to complex environments. They often rely on predefined distributions, static rules, or fixed feature representations, making it difficult to capture evolving characteristics such as dynamic item departures, contextual heterogeneity, and temporal nonstationarity. Moreover, the performance of these methods heavily depends on manually designed rules and expert-driven feature capturing. Furthermore, varying contextual patterns necessitate distinct heuristics, rendering a one-size-fits-all approach impractical in dynamic environments. Designing and maintaining these rules requires substantial human expertise and extensive exploration effort, and experts may face substantial challenges in timely adapting decision logic as contextual patterns evolve. As a result, it remains challenging to develop an adaptive framework that can generalize across diverse dynamic contexts.

Recent advances in large language models (LLMs) offer a promising direction to overcome these challenges.
LLMs have been applied in OM, demonstrating their potential in problem modeling and decision support \citep{huang2025orlm, bertsimas2024robust, wang2024survey}. Unlike human experts constrained by cognitive biases and a limited search space, LLMs leverage pre-trained knowledge to explore infinite-dimensional algorithmic spaces \citep{cao2024survey}.
Leveraging their powerful reasoning and code-generation capabilities, LLMs can autonomously identify relevant features, synthesize heuristic rules, and iteratively refine decision logic \citep{yang2024mindllm, li2025fundamental}. Such capabilities position LLMs as a new paradigm for adaptive optimization in dynamic scheduling environments \citep{tang2025llm, cheng2025large}. Nevertheless, existing LLM-based approaches remain fundamentally limited when applied directly to complex, nonstationary planning problems \citep{kambhampati2024llms}. Our motivating experiments also reveal that a directly LLM-based heuristic design is not suitable for dynamic environments; detailed results are shown in Table \ref{tab:single-heu}.


\begin{figure}[htbp]
    \centering
    \caption{The Overview of Context-Aware Scheduler - MiCo.}
    \label{fig: overview}
    \vspace{2mm}
    \includegraphics[width=1\textwidth]{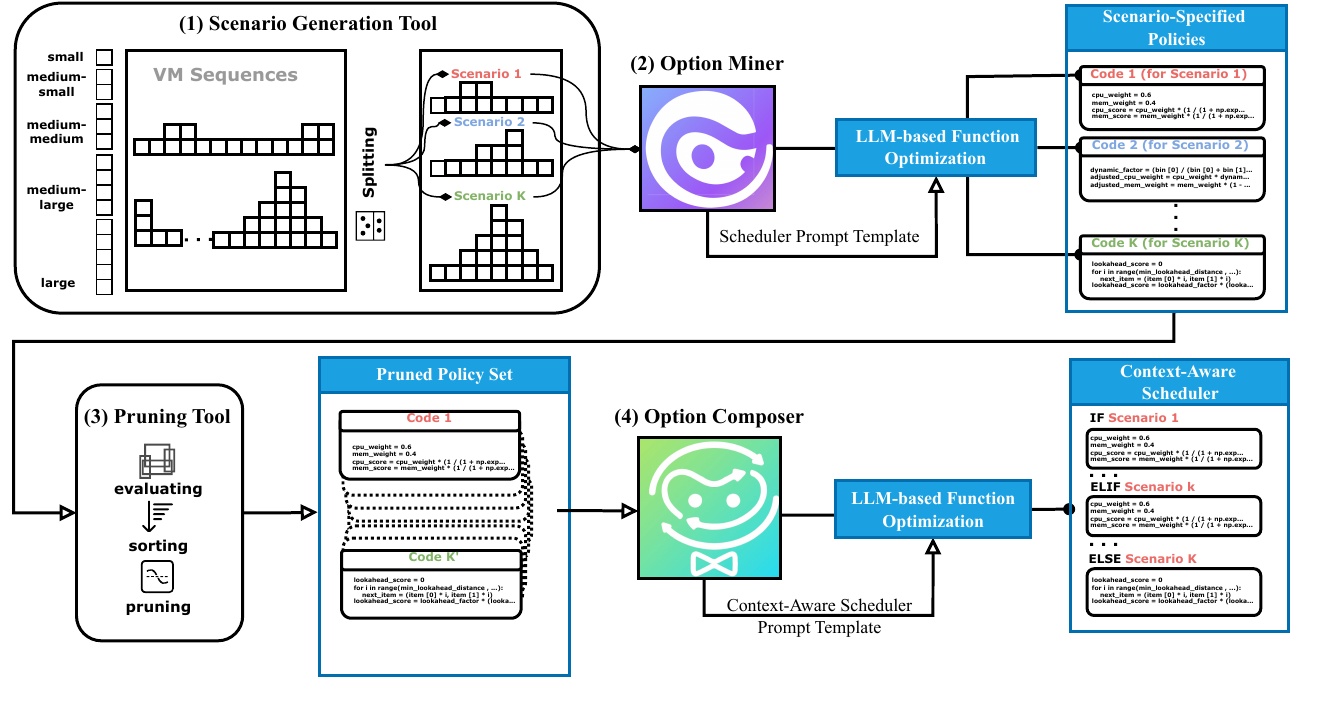}
\end{figure}

\subsection{Our Framework and Contributions}

To bridge this gap, we propose MiCo, a hierarchical language agent framework, for automated VM scheduling that replaces traditional heuristic designs in the ODMBP problem.
MiCo is conceptually grounded in the Semi-Markov Decision Process with Options (SMDP-Option) framework~\citep{sutton1999between, baykal2010semi}, which provides temporal abstraction for hierarchical decision-making.
MiCo consists of two language agents: the \textit{\textbf{Option Miner}} and the \textit{\textbf{Option Composer}}, which correspond to two fundamental learning processes in the SMDP-Option framework~\citep{sutton1999between,li2023hit}: automated option discovery and master policy learning. 
As Figure~\ref{fig: overview} illustrates, in the automated option discovery phase, a scenario-generation tool first partitions the raw VM request streams into contextually coherent temporal slices, which are defined as scenarios. For each scenario, the Miner emulates domain experts by designing candidate scheduling strategies via LLM-based function optimization, from which it derives scenario-specific policies. 
This phase operates within relatively stable environments, allowing the Miner to fully explore the heuristic search space without requiring online contextual adaptation. Together, these scenarios and their associated policies form the option set. In the master policy learning phase, a pruning tool is applied to eliminate redundancy, and then the Composer employs LLM-based function optimization to develop a context-aware scheduler from the pruned policy set. The Composer explicitly leverages contextual workload patterns to determine which policy should be activated under the current system state, thereby adapting to the nonstationary nature of the workload. This scheduler dynamically selects among the pruned policies based on historical VM patterns (see Sections~\ref{sec: 3.3} and \ref{sec: 3.4} for details).


We validate the effectiveness of our framework through comprehensive experiments on a real-world VM dataset. 
The results demonstrate that our heuristic approach generated by LLMs consistently outperforms established baselines such as Best-Fit, First-Fit, Hindsight, and SchedRL. 
Across diverse scenarios, the proposed hierarchical architecture exhibits strong robustness, achieving near-optimal performance and adapting effectively to dynamic demand patterns. Beyond performance gains, the heuristics discovered by the LLM display meaningful structural similarities to classical optimization strategies, which underscores both their interpretability and their potential to enrich domain expertise.

This paper advances online bin packing for cloud computing resource management through three key contributions:
1) \textit{LLM-Driven Heuristic Framework for ODMBP}: We formulate VM scheduling as an ODMBP problem and pioneer an LLM-driven heuristic design paradigm. 
The framework automatically discovers interpretable, context-contingent scheduling rules, thereby reducing the substantial exploratory effort typically required from domain experts.
2) \textit{Hierarchical Architecture for Contextual Adaptation}: We develop MiCo, a hierarchical language-agent architecture grounded in the SMDP-Option framework. This design enables systematic scalability and robust adaptation to nonstationary request patterns, outperforming heuristic baselines and RL approaches, achieving 96.9\% average performance ratio.
3) \textit{Open-source Implementation for LLM-Based Combinatorial Optimization.}
We release an open-source implementation supporting extendable language-based heuristic optimization. The framework is designed for researchers and practitioners to apply the paradigm to other combinatorial optimization problems, facilitating reproducibility and accelerating methodological innovation in operations.

The remainder of this paper is organized as follows. Section \ref{sec: sec2} formulates the VM scheduling problem as a sequential decision process and examines direct LLM-based scheduling methods, whose limitations motivate a hierarchical solution. Section \ref{sec: method} introduces MiCo, the proposed hierarchical language-agent framework based on the SMDP-Option formulation, and details the Option Miner and Option Composer. The complete experimental results are demonstrated in Section \ref{sec: experiments}. Finally, the paper is concluded in Section \ref{sec: conclusion}.

\section{Problem Formulation and Motivating Experiments} \label{sec: sec2}
In Section \ref{sec: problem}, we develop the ODMBP and MDP formulations of the VM scheduling problem and specify the sequential optimization objective. In Section \ref{sec: 3.1}, we then analyze direct LLM-based optimization of this objective, and subsequently refine it with a context-aware methodology.

\subsection{Problem Formulation} \label{sec: problem}
As illustrated in Figure \ref{fig: sched illu}, the VM scheduling problem can be naturally framed as an ODMBP problem. The system consists of a set of $N$ PMs, each represented as a bin with a $d$-dimensional remaining capacity vector $\boldsymbol{c}_i^{pm} \in \mathbb{Z}^d$, where $i \in \{1,\dots, N\}$ and $d$ denotes the number of resource dimensions (e.g., CPU, memory, storage). Each arriving VM request is treated as an exogenous input $w=(vm, \boldsymbol{c}^{vm}, b)$, where $vm$ denotes the request ID, $\boldsymbol{c}^{vm} \in \mathbb{Z}^d$ denotes the demand. $b \in \{0,1\}$ indicates the operation type ($1=$ creation, $0=$ deletion).


We model VM scheduling in an event-driven process over decision steps \(t=1,\dots,T\). At step \(t\), the observation is \((\{\boldsymbol c^{pm}_{i,t}\}_{i=1}^N, w_t)\). If \(b_t=1\), the scheduler choose \(a_t\in\{0,1,\dots,N\}\) with feasibility \(\boldsymbol c^{pm}_{a_t,t}\ge \boldsymbol c^{vm}_t\) (or reject \(a_t=0\)). Once a VM is allocated, it cannot be migrated before its departure. To track placements, we maintain an allocation map $\Phi_t$ that maps each active VM to its hosting PM, and a deletion queue $\mathcal Q_t$ that stores VMs whose deletion events have arrived but whose resources are released only when the next creation event is processed (synchronizing state updates with allocation decisions). The endogenous system state is thus
$\big(t,\{\boldsymbol c^{pm}_{i}\}_{i=1}^N,\Phi,\mathcal Q\big)$.
Furthermore, we model the VM scheduling problem as a finite-horizon, discrete-time MDP with Exogenous Inputs \citep{sinclair2023hindsight}, which is characterized by the tuple \( \langle \mathcal{S}, \mathcal{A}, \mathcal{P}, \mathcal{R} \rangle \), to capture the online sequential decision-making process.

\noindent{-} \textbf{State Space ($\mathcal{S}$)}: The state $s_t \in \mathcal{S}$ at time $t$ integrates both endogenous system state and exogenous input:
\begin{equation*}
s_t = \Big( t, \{\boldsymbol{c}^{pm}_{i,t}\}_{i=1}^N, 
\Phi_t, \mathcal{Q}_t, 
w_t  \Big).
\end{equation*}

\noindent{-} \textbf{Action Space ($\mathcal{A}$)}: The action space is $\mathcal{A}=\{\varnothing, 0,1,\cdots, N\}$, where $a_t=0$ indicates reject the current request. The feasible placement set at $s_t$ is
$\mathcal A(s_t)=\{i\in\{1,\dots,N\}:\boldsymbol c^{pm}_{i,t}\ge \boldsymbol c^{vm}_t\}\cup\{0\}$.
When $b_t=0$ (deletion), no explicit decision is required and we take a dummy action $a_t=\varnothing$.


\noindent{-} \textbf{Reward ($\mathcal{R}$)}: The reward function $r(\cdot, \cdot)$ is designed to reflect successful allocations. The reward function assigns $r(s_t, a_t) = 1$ for successful placements $a_t \neq 0$ and $0$ otherwise. We adopt a first-failure termination rule: if at any creation step no feasible PM is found or the request is rejected, the process enters an absorbing terminal state and yields zero reward thereafter.
\begin{remark}
A more general reward formulation can incorporate additional terms reflecting resource utilization or workload heterogeneity, such as  
$r\big(s_t,a_t\big) = \mathbf{1}\{a_t \neq 0\} + \alpha \cdot \text{Util}\big(c^{pm}_{i,t}\big) + \beta \cdot \text{Type}\big(c^{vm}_t\big)$,  
where $\text{Util}(\cdot)$ denotes PM resource balance and $\text{Type}(\cdot)$ scales with VM size.  
This general form allows flexible extensions to other scheduling objectives.
\end{remark}

\noindent{-} \textbf{Transitions ($\mathcal{P}$)}: The transition $\mathcal{P}(s_t \mid s_{t-1},a_{t-1})$ updates occur at request arrivals. For deletion events ($b_t=0$), the system only enqueues the VM for delayed release:
$\mathcal Q_{t^+}=\mathcal Q_t\cup\{vm_t\}$,
while PM capacities are unchanged at that step.
For creation events ($b_t=1$), before executing the placement action, we update the deletion queue: for each PM $j$, we add back the total demand of queued VMs that were previously hosted on $i$,
\[
\forall i:\quad
\boldsymbol c^{pm}_{i,t}\leftarrow \boldsymbol c^{pm}_{i,t} + \sum_{v\in \mathcal Q_t}\boldsymbol c^{vm}_v\,\mathbf 1\{\Phi_t(v)=i\},
\]
and let $\mathrm{dom}(\Phi_t)$ denote the set of active VMs currently mapped by $\Phi_t$. We update $\Phi_t \leftarrow \Phi_t\big|_{\mathrm{dom}(\Phi_t)\setminus \mathcal{Q}_t}$ and reset $\mathcal Q_t\leftarrow \emptyset$.
Then, if $a_t=i\neq 0$ is feasible (i.e., creation request), we allocate the VM by
$\boldsymbol c^{pm}_{i,t+1}=\boldsymbol c^{pm}_{i,t}-\boldsymbol c^{vm}_t$ and set
$\Phi_{t+1}(vm_t)=i$ and $\Phi_{t+1}(v)=\Phi_t(v)$ for all $v\in\mathrm{dom}(\Phi_t)$; otherwise we transition to the absorbing terminal state.

\noindent{-} \textbf{Objective Function:} The objective is to find a  scheduling policy $\pi$ that maximizes the expected undiscounted cumulative reward:
\begin{equation} \label{eq: mdp-obj}
\max_\pi \mathcal{J}_\pi(s_t) =\mathbb{E}_{\pi, \mathcal{P}} \left[\textstyle{\sum_{k=t}^T r\left(s_{k}, a_{k}\right)} \Big| s_t \right].
\end{equation} 

\subsection{Motivating Experiment: Context-Aware Function Optimization with LLM Agents}\label{sec: 3.1}
\subsubsection{LLM-based Function Optimization} \label{sec: llm-based-fn}
To optimize the scheduling objective in Eq. \eqref{eq: mdp-obj}, we characterize a policy $\pi$ as a function code representation and leverage LLMs for automated code generation and algorithmic optimization. The language agent implements policy improvement through structured code evolution, as expressed in Eq. \eqref{eq:llm_des_origin}. We enhance the LLM reasoning process through an improved variant of \textit{contrastive prompting} \citep{chia2023contrastive, li2024learning}. Instead of presenting correct\/incorrect pairs, our approach selects the top-$M$ high-performing policies $\{\pi^{(m)}\}_{m=1}^M$ from historical executions. Upon receiving a prompt that includes top-$M$ policies $\{\pi^{(m)}\}_{m=1}^M$ encapsulated in code form, along with a role description and a task description, a language agent is capable of generating a refined and presumably more optimal policy \(\pi'\), which is also represented in code form. 

\begin{equation}
    \pi' = \mathcal{LLM}\left(\text{role\_des}, \text{task\_des},\{\pi^{(m)}\}_{m=1}^M, \xi \right),
    \label{eq:llm_des_origin}
\end{equation}
where $\xi$ denotes the sampling stochasticity that controls the exploration of LLM, i.e., the temperature.

\begin{figure}[htbp]
    \centering
    \caption{The Framework of LLM-based Function Optimization.}
    \label{fig: LLM-fn-opt}
    \vspace{2mm}
    \includegraphics[width=1\textwidth]{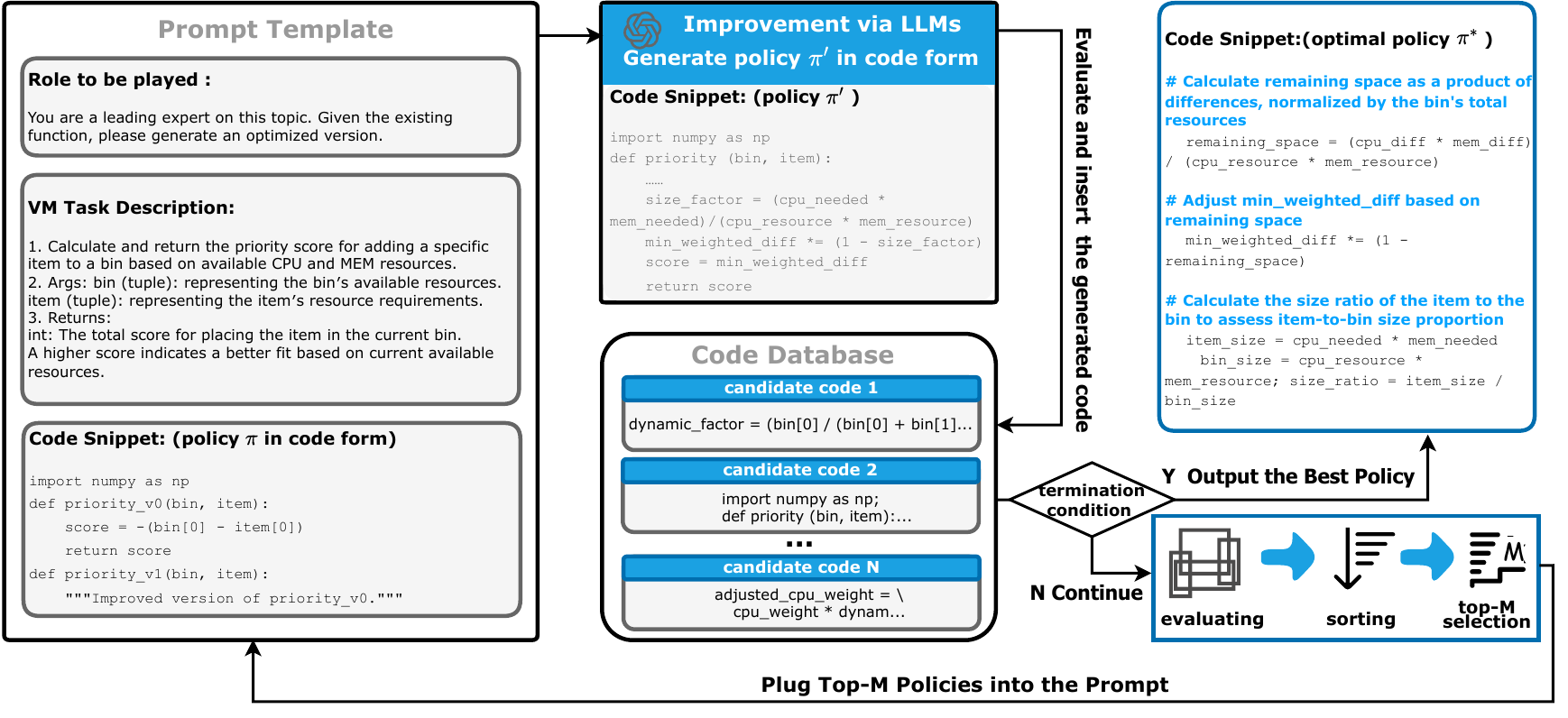}
\end{figure}

The framework operates through a closed-loop optimization process as shown in Figure \ref{fig: LLM-fn-opt}, maintaining a dynamic set of top-$M$ candidate policies at each iteration. The process begins with an initial policy \(\pi^{(0)}\), provided as code along with a role description and a task description. 
During each iteration, the language agent generates enhanced candidate policies $\{\pi'\}$ through contrastive analysis of the current top-$M$ policies $\{\pi^{(m)}\}_{m=1}^M$ (selected by $\mathcal{J}_\pi$ ranking). Then, the next-generation population is formed by selecting the updated top-$M$ performers from the union of historical and newly generated policies.
The evolutionary process continues iteratively until a termination condition is met. 


\subsubsection{Context-Aware Policy Learning} \label{sec: context-aware}
We next examine whether LLM-based scheduling policies can be directly optimized without explicitly distinguishing workload scenarios. The experiments in this subsection use the VM scheduling dataset with implementation details and environment specifications provided in Section \ref{sec: setup}. We evaluate all methods using the performance ratio defined in Eq. \eqref{eq: per-ratio}. To conduct direct policy optimization, we adopt FunSearch as the LLM-driven heuristic evolution framework and Best-Fit as the heuristic baseline. FunSearch iteratively refines scheduling rules through large language model reasoning and performance-based feedback, enabling automatic discovery of interpretable heuristic structures \citep{romera2024mathematical}.

Since this experiment removes scenario generation and contextual characterization, we regard it as an ablation study of scenario-aware decomposition (as detailed in Section~\ref{sec: ablation study}), and its quantitative results are therefore reported in Table \ref{tab:single-heu}. Here, we focus on the methodological insight derived from this evidence.
When request patterns vary substantially over time, policies learned under \textit{MiCo w/o Scenario \& Context}, which correspond to a single static heuristic without contextual inputs, exhibit unstable performance. Such policies tend to align with frequent or dominant request regimes, but their effectiveness deteriorates as the workload distribution shifts. To address this limitation, we introduce contextual information and consider \textit{MiCo w/o Scenario}, where a single context-aware policy is trained across all sequences. However, this design still fails to deliver consistent performance: while it adapts to short-term fluctuations, it struggles to capture broader nonstationary structures. In contrast, training context-aware policies separately for individual request traces, denoted as \textit{MiCo w/o Scenario (Per-Trace)}, yields strong specialization. Yet, this one-to-one specialization severely limits transferability and scalability, constraining practical deployment.

In conclusion, these findings demonstrate that neither purely context-independent nor purely context-aware designs can reconcile the trade-off between specialization and generalization. This observation motivates the hierarchical design developed in Section \ref{sec: method}, where context-independent policy discovery is performed within scenarios to extract stable heuristic structures, and a context-aware master policy composes these heuristics dynamically to enable adaptive scheduling under nonstationary workloads.

\section{The MiCo  Framework} \label{sec: method}

We address context dependence and nonstationarity by reformulating the VM scheduling problem as an SMDP-Option problem in Section \ref{sec: 3.2}. The resulting hierarchical structure separates option discovery at the micro-level, handled by the Option Miner (in Section \ref{sec: 3.3}), and option planning at the macro-level, handled by the Option Composer (in Section \ref{sec: 3.4}). Section \ref{sec: 3.5} summarizes the full algorithmic procedure.

\subsection{Problem Reformulation: SMDP-Option} \label{sec: 3.2}
\begin{figure}[htbp]
    \centering
    \caption{Diagram of MDP, SMDP, and Options over MDP.}
    \label{fig: option}
    \includegraphics[width=0.8\textwidth]{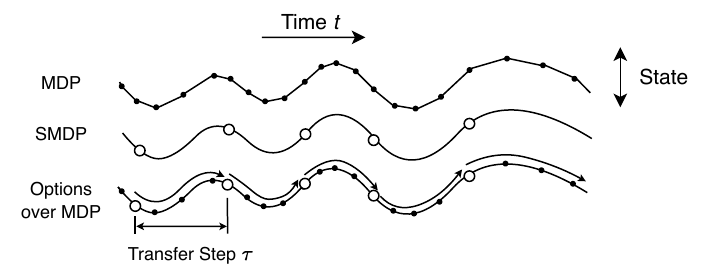}\\
    \FigureNoteStylere{In an MDP, the progression occurs through small, discrete-time steps, in contrast to an SMDP, which features larger, continuous-time transitions. The option constitutes a temporally abstract action that executes over several time steps, allowing the trajectory of an MDP to be analyzed at either of the detailed, discrete-time level or the more abstract, continuous-time level.}
\end{figure}

Given the observed limitations under nonstationary contexts in Section \ref{sec: context-aware}, we reformulate the problem as an SMDP-Option to represent it in a multi-time scale manner, as illustrated in Figure \ref{fig: option} (cf. Figure 1 in \citet{sutton1999between}). This hierarchical decomposition simplifies computations by abstracting sequences of actions into macro-level decision-making units.

The term \textit{options} generalizes primitive actions to include temporally extended courses of action. In this work, we adopt Markov options, where both the intra-option policy and termination condition depend only on the current environment state. In contrast, the macro-level option selection policy may incorporate finite history information at option switching times. Thus, the micro-level execution remains a standard MDP, while the macro-level decision process becomes an SMDP. We further define the options:

\begin{definition}[Option]
    An option $o$ is a tuple $\langle \mathcal{I}, \pi, \beta \rangle$. \textit{Input Set}: $\mathcal{I} \subseteq \mathcal{S}$ specifies the states where the option can be initiated. In our setting, $\mathcal{I} = \mathcal{S}$.
    \textit{Intra-Option Policy} $\pi: \mathcal{S} \times \mathcal{A} \rightarrow [0,1]$ determines action selection while the option is active. \textit{Termination Condition} $\beta: \mathcal{S} \rightarrow [0,1]$ specifies the probability of terminating the option in state $s$: (i) resource-based termination when no PM has sufficient remaining capacity for the incoming VM, i.e., $\boldsymbol{c}^{pm}_i < \boldsymbol{c}^{vm}, \forall i $; and (ii) a horizon limit $\tau_{max}$, such that the option duration $\tau(o)$ is a truncated geometric random variable.
\end{definition}

At option switching times, the master policy may utilize a finite-length history context $h_t^{L}=\psi(s_{t-L:t})$, where $\psi(\cdot)$ is a fixed feature extractor over the recent $L$ states (or requests). We denote the corresponding augmented state space by $\mathcal{H}$. When $L=0$, $h_t^{0}$ reduces to the standard Markov state $s_t$. Once an option $o\in\mathcal{O}$ is selected at a switching time with macro state $h_t^{L}$, the environment evolves for $\tau(o)$ steps under $(\pi,\beta)$ and terminates in  $s_{t+\tau(o)}$, inducing the next macro state $h_{t+\tau(o)}^{L}$. This defines a Semi-MDP transition kernel $\mathcal{P}'(h',\tau \mid h,o)$, and expected cumulative reward, $\mathcal{R}'(h,o)$.
\begin{proposition}[\cite{sutton1999between}]
    \label{proposition}
    Given any MDP $\mathcal{M} = \langle \mathcal{S}, \mathcal{A}, \mathcal{P}, \mathcal{R} \rangle$ and option set $\mathcal{O}$, the decision process that selects options based on the augmented state $\mathcal{H}$ forms a Semi-MDP $\mathcal{M}' = \langle \mathcal{H}, \mathcal{O}, \mathcal{P}', \mathcal{R}' \rangle$.
\end{proposition}

As formally established by Proposition \ref{proposition}, this hierarchical formulation preserves the Markov property at the micro level, since each option acts on the original MDP state space. Meanwhile, it introduces temporal abstraction at the macro level by allowing decisions to span multiple time steps. In VM scheduling, an option corresponds to a coordinated multi-step scheduling strategy that continues until termination. Hence, the underlying MDP remains the same; only the time granularity of decision-making changes.


At the micro level, the intra-option policy \( \pi_o\) dictates the scheduling actions \( a_t \sim \pi_o(\cdot \mid s_t) \) to maximize the cumulative rewards until the option \( o \) terminates in $s_{t+\tau(o)}$. 
Given an initiation at primitive state $s_t$ and an option $o$, the intra-option return is the cumulative reward accrued during the option execution:

\begin{equation}\label{eq: micro-reward}
\max_{\pi_o} \mathcal{J}_{\pi_o}(s_t, o) = \mathbb{E}_{\pi_o, \mathcal{P}}\!\left[
\sum_{m=0}^{\tau(o)-1} \, r\big(s_{t+m},a_{t+m}\big)
\;\Big|\; s_t, o
\right].
\end{equation}

Eq. \eqref{eq: micro-reward} is a conditional and truncated version of the original MDP objective in Eq. \eqref{eq: mdp-obj}, restricted to the execution horizon of option $o$. Unlike the global unrestricted policy in Eq. \eqref{eq: mdp-obj}, the intra-option policy $\pi_o$ only governs actions while option $o$ is active.

At the macro level, when initiated in a state $s_t$, the master policy over options $\mu$ selects an option $o_t \in \mathcal{O}_s$ such that \( o_t \sim \mu(\cdot \mid h_t) \). The objective is to maximize the overall reward:
\begin{equation} \label{eq: macro-reward}
\max_\mu \mathcal{J}_\mu(h_t)=\mathbb{E}_{\mu, \mathcal{P}'}\!\Big[ 
\; \mathcal{J}_{\pi^*_o}(s_t, o_t) \;+\; 
\mathbb{E}_{\tau(o_t),\, \mathcal{P}}\big[ \mathcal{J}_\mu(h_{t+\tau(o_t)}) \mid s_t,o_t \big]
\Big].
\end{equation}


Eq. \eqref{eq: macro-reward} uses the optimal intra-option value $\mathcal{J}_{\pi^*_o}(s_t, o_t)$ from Eq. \eqref{eq: micro-reward} as the immediate reward of each macro-level decision. Thus, the macro-level SMDP compresses multiple primitive MDP steps into a single high-level transition whose duration is $\tau(o_t)$. In MiCo, options are mined offline and then treated as fixed behaviors; the composer optimizes $\mu$ to select among them under nonstationarity.

\subsection{Options Discovery: Context-Specific Option \textbf{Mi}ner} \label{sec: 3.3}

The option discovery stage aims to find the scheduling policies that can be composed to obtain good performance in scheduling. 
As illustrated in Figure \ref{fig: overview},  this process combines \textit{scenario generation} and \textit{language-agent guided option learning}. First, a scenario generation tool partitions the VM request stream into $K$ contextual segments, each characterizing distinct demand patterns. Policy learning subsequently operates within these scenarios, enabling the focused optimization of scenario-specific policies. This approach addresses the challenge of option evaluation under nonstationary constraints by ensuring policy adaptation to individual scenario characteristics.

\paragraph{Scenario Generation.}
Given a request sequence $\{w_t\}_{t=1}^T$ that covers the entire planning horizon of length $T$, the sequence is partitioned into $K$ contiguous, non-overlapping scenarios. Specifically, given a window length $W$ (in time units), the timeline is divided into $K = \lceil T / W \rceil$ intervals, each corresponding to a scenario $S_k$: 
\begin{equation}
\label{eq: scenario gen.}
    S_k = \{ w_t:\; (k-1)W < t \le \min(kW,T)\},\quad k=1,\dots,K.
\end{equation}

\begin{remark}
The scenario generation provides approximately stationary slices for stable option mining. To further assess robustness, we additionally study a data-driven scenario generation scheme that infers scenario boundaries from empirical patterns, with detailed results reported in Appendix~\ref{secA2}, Figure EC.4. We intentionally keep the scenario construction lightweight and low-assumption, requiring minimal expert knowledge. This keeps the setup LLM-executable and avoids injecting domain heuristics that could confound the subsequent optimization.
\end{remark}

\paragraph{Language-Agent Guided Option Learning.}
Once the scenarios have been generated, the Option Miner employs an LLM-based function optimization approach (Section \ref{sec: llm-based-fn}) to learn options within scenarios. The learning process follows an iterative evaluation-improvement loop to refine scheduling strategies for nonstationary VM request scenarios.  

\noindent - Policy Evaluation Phase: From each scenario $S_k$, we first randomly sample \( n_{s} \) representative task sequences. Then, for each candidate policy \( \pi_{o_k}^{(n)} \) in scenario \( S_k \), we evaluate performance \( \mathcal{J}_\pi \) as the average reward under Eq. \eqref{eq: micro-reward}  across these \( n_{s} \) sequences.  

\noindent - Policy Improvement Phase:  We first replace the template prompt with the scheduler template (Appendix \ref{box:1}) and encode the top-$M$ performing policies $\big\{\pi_{o_k}^{(n,m)}\big\}_{m=1}^M$ (ranked by $\mathcal{J}_\pi$) into the prompt. Next, improved candidate policies are generated via LLM reasoning as follows. Finally, the top-$M$ candidate policies achieving the highest \( \mathcal{J}_\pi \) (i.e., Eq. \eqref{eq: micro-reward}) are retained for the next iteration. 
\begin{equation} \label{eq:miner policy}
    \pi_{o_k}^{(n+1)} = \mathcal{LLM}\left(\text{role\_des}, \text{task\_des}, \big\{\pi_{o_k}^{(n,m)}(\text{item}^{vm}, \text{bin}^{pm})\big\}_{m=1}^M, \xi \right).
\end{equation}

The loop iteratively refines option policies until a budget is reached. The final iterated policies form a library of options $\mathcal{O}=\{o_k\}_{k=1}^K$ with intra-option policies $\{\pi_{o_k}\}$.

\subsection{Planning over Options: Option Composer} \label{sec: 3.4}
Before composing policies, we conduct \textit{option pruning} to reduce policy search space. The Option Composer establishes a two-tier decision hierarchy to coordinate the discovered options in nonstationary environments. At the macro level, it maintains a master policy $\mu$ that selects options $o$ based on real-time context features, while activated options autonomously handle micro-level resource allocation through their intra-policies $\pi_o$. 

\paragraph{Option Pruning.}
To further reduce the search space and enhance policy efficiency, the policy set is pruned according to a heuristic filtering criterion. A policy $\pi_k$ is retained if it satisfies:  
(i) near-optimal performance in its own scenario $S_k$ (\textit{single-scenario excellence}),  
and  (ii) above-average performance in a sufficient fraction of all scenarios (\textit{cross-scenario robustness}).

\begin{equation}
\label{eq: pruning}
\mathcal{O}^{'}
=\left\{
  o_k \;\middle|\;
    \mathcal{J}_{\pi_{o_k}} (S_k) \geq q_1 \cdot \mathcal{J}^{*}(S_k)
  \;\;\wedge\;\;
    \frac{1}{K}\sum_{j=1}^{K}
      \mathbf{1}\!\Bigl\{
        \mathcal{J}_{\pi_{o_k}}(S_j)
        \;\ge\; q_2 \cdot 
        \bar{\mathcal{J}}(S_j)
      \Bigr\}
    \;\ge\;q_3
\right\}.
\end{equation}

Here, $\mathcal{J}^*(S_k)=\max_{o}\mathcal{J}_{\pi_o}(S_k)$ denotes the best achievable performance in scenario $S_k$, and $\bar{\mathcal{J}}(S_j)=\frac{1}{K}\sum_{k=1}^K \mathcal{J}_{\pi_{o_k}}(S_j)$ is the scenario average performance across all candidate policies. The pruning rule is controlled by parameters $q_1$, $q_2$, and $q_3$. Guided by preliminary analysis, we set $q_1 = q_2 = 0.95$ to ensure stringent per-scenario quality, and $q_3 = 0.5$ so that a policy is retained only if it outperforms the scenario average in at least half of the scenarios.

\paragraph{Language-Agent Guided Master Policy Learning.} The macro-level policy \( \mu \) selects the optimal option \( o \in \mathcal{O}' \) based on the system state at option switching times. The context-aware composer learns a master policy \( \mu \) through the same LLM-based function optimization (Section \ref{sec: llm-based-fn}). Unlike the context-independent strategy optimization of Option Miner, the Option Composer incorporates historical context $h_t^L=\psi(s_{t-L:t})$ to coordinate scheduling strategies across scenarios.

\noindent - Policy Evaluation Phase:
From the full request stream $\{w_t\}_{t=1}^T$, we randomly sample \( n_{s} \) task sequences covering diverse exogenous system states. Then, for the master policy \( \mu^{(n)} \) at iteration \( n \), we evaluate coordination performance $\mathcal{J}_\mu$ from Eq. \eqref{eq: macro-reward}, calculated as the average reward across all \( n_{s} \) sequences.
    
\noindent - Policy Improvement Phase: Replace the template prompt with the context-aware scheduler template (Appendix \ref{box:2}) and encode the top-$M$ policies \( \big\{\mu_{o_k}^{(n,m)} \big\} _{m=1}^M \) into the prompt. Then, generate improved candidate policies via LLM reasoning as follows.  Finally, retain the policy achieving the highest \( \mathcal{J}_\mu \) for the next iteration. 
\begin{equation} \label{eq:composer policy}
    \mu^{(n+1)} = \mathcal{LLM}\left(\text{role\_des}, \text{task\_des},  \big\{\mu^{(n,m)}(\text{item}^{vm}, \text{bin}^{pm}, \text{context}^{vm}) \big\}_{m=1}^M,  \xi \right).
\end{equation}

The Option Composer iteratively improves the master policy \( \mu \), searching for the best strategy to maximize Eq. \eqref{eq: macro-reward}. The loop repeats until convergence or a budget is reached, and the best-performing master policy is selected as $\mu^*$.

\subsection{Execution Stage}\label{sec: 3.5}

\begin{algorithm}[t]
\caption{Hierarchical LLM Agent for VM Scheduling}
\label{alg:hierarchical}
\begingroup  
\linespread{1.2}\selectfont  
\begin{algorithmic}[1]
\State \textbf{Input:} VM request sequence $\{w_t\}_{i=1}^T$, PM cluster $\{\boldsymbol{c}^{pm}_{i}\}_{i=1}^N$, and algorithm parameters illustrated in Table \ref{tab:parameter_list}.

\State \textbf{Output:} Context-aware scheduler $\mu^*$
\State Split requests $\{w_t\}_{t=1}^T$ into $\{S_k\}_{k=1}^K$  using Scenario Generation Tool Eq. \eqref{eq: scenario gen.}
\Function{\textcolor{myblue}{OptionMiner}}{$\{S_k\}^K_{k=1},\{\boldsymbol{c}^{pm}_{i}\}_{i=1}^N$}
    \For{$k=1$ to $K$} 
        \State Initialize $\pi_{o_k}^{(0)} \gets \mathrm{SeedPolicy}(S_k)$; history $\mathcal{D}_k \gets \emptyset$
        \For{$n=0$ to $M_{Mi}-1$}
            \State $\mathcal{J}_{o_k}^{(n)} \gets \mathrm{Evaluate}(\pi_{o_k}^{(n)}, S_k)$ using \text{Eq.} \eqref{eq: micro-reward}
            \State $\mathcal{D}_k \gets \mathcal{D}_k \cup \{(\pi_{o_k}^{(n)}, \mathcal{J}_{o_k}^{(n)})\}$
            \State $\pi_{o_k}^{(n+1)} \gets \mathrm{UpdateByMiner}(\mathcal{D}_k, \text{Eq.\eqref{eq:miner policy}})$ 
        \EndFor
        \State $\pi^*_{o_k} \gets \arg\max_{\pi} \mathcal{J}_\pi$ from $\mathcal{D}_k$
    \EndFor
    \State \Return $\mathcal{O}=\{o_k\}_{k=1}^{K}$ with intra-option policies $\{\pi_{o_k}^{*}\}_{k=1}^{K}$
\EndFunction
\State $\mathcal{O}\gets$ \Call{\textcolor{myblue}{OptionMiner}}{$\{S_k\}_{k=1}^K$}
\State
      $\mathcal O'\gets \mathrm{UpdateByPruning} (\mathcal{O}, \text{Eq.} \eqref{eq: pruning})
      $ 
\Function{\textcolor{myblue}{OptionComposer}}{$\mathcal O',\{w_t\}^T_{t=1},\{\boldsymbol{c}^{pm}_{i}\}_{i=1}^N$}
    \State Initialize $\mu^{(0)} \gets \mathrm{SeedScheduler}(\mathcal{O}')$; history $\mathcal{D}_\mu \gets \emptyset$
    \For{$n=0$ to $M_{Co}-1$}
        \State $\mathcal{J}_{\mu}^{(n)} \gets \mathrm{Evaluate}(\mu^{(n)}, \{w_t\}^T_{t=1})$ using \text{Eq.} \eqref{eq: macro-reward}
        \State $\mathcal{D}_\mu \gets \mathcal{D}_\mu \cup \{(\mu^{(n)}, \mathcal{J}_{\mu}^{(n)})\}$
        \State $\mu^{(n+1)} \gets \mathrm{UpdateByComposer}(\mathcal{D}_\mu, \text{Eq.\eqref{eq:composer policy}})$ 
    \EndFor
    \State \Return $\mu^* \gets \arg\max_{\mu} \mathcal{J}_{\mu}$ from $\mathcal{D}_\mu$
\EndFunction
\State $\mu^* \gets$ \Call{\textcolor{myblue}{OptionComposer}}{$\mathcal{O}'$}
\end{algorithmic}
\endgroup
\end{algorithm}

The hierarchical algorithm framework is illustrated in Algorithm \ref{alg:hierarchical}. Once the optimal scheduling policy $\mu$ has been learned, the execution process begins with initial state \( s_0 \) and iteratively performs option selection and execution:

\begin{enumerate}
\item \textbf{Option Selection:} At the $k$-th decision point \( t_k \), the composer observes current state \( s_{t_k} \) and selects option \( o_k \sim \mu^*(\cdot |h^L_{t_k}) \) from \( \mathcal{O}' \).
\item \textbf{Option Execution:} While option $o_k$ is active, the agent repeatedly samples primitive actions according to the intra-option policy,
\[
a_{t_k+m} \sim \pi_{o_k}(\cdot \mid s_{t_k+m}), \qquad m=0,1,\dots,\tau_k-1,
\]
until the option terminates after a (random) duration $\tau_k$ according to its termination rule $\beta_{o_k}$ (e.g., triggered by resource-violation handling or a timeout).
\item \textbf{State Transition:} Upon termination, control returns to the master policy at the next decision point $t_{k+1}=t_k+\tau_k$ with the updated state $s_{t_{k+1}}$ (distributed according to the induced SMDP kernel $\mathcal{P}'(\cdot\mid h^L_{t_k},o_k)$).
\end{enumerate}


\section{Experiments}\label{sec: experiments}
In this section, we evaluate the performance of our proposed algorithm for VM scheduling. 
Section \ref{sec: setup} outlines the experimental setup, and Section \ref{sec: results} presents the main performance results, ablation study and robustness analysis. Finally, Section \ref{sec: interpretable insights} investigates whether the strategies discovered by the large language model align with traditional heuristic approaches.



\subsection{Experimental Setup}\label{sec: setup}

\begin{figure}[htbp]
    \centering
    \caption{Characteristics of VM Types and Scenarios in Huawei Dataset.}
    \label{fig:vmtype-scenario}
    \vspace{-4mm}
    \begin{subfigure}[t]{0.4\textwidth}
        \caption{\small}
        \label{fig:4.1sub1}
        \vspace{-4mm}
        \includegraphics[width=\textwidth]{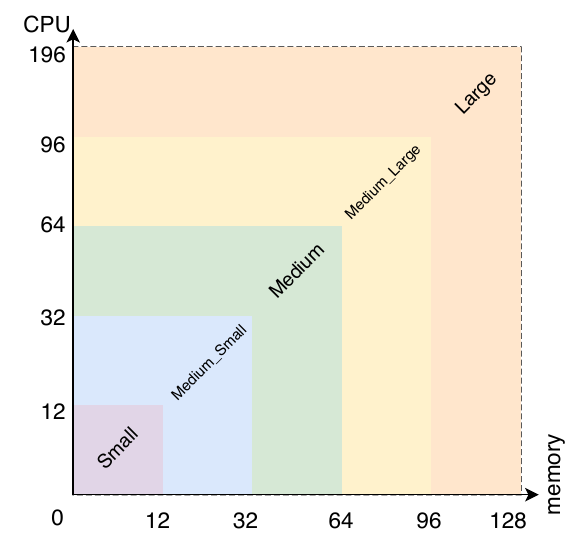}
    \end{subfigure}
    \begin{subfigure}[t]{0.57\textwidth}
        \caption{\small }
        \label{fig:4.1sub2}
        \includegraphics[width=\textwidth]{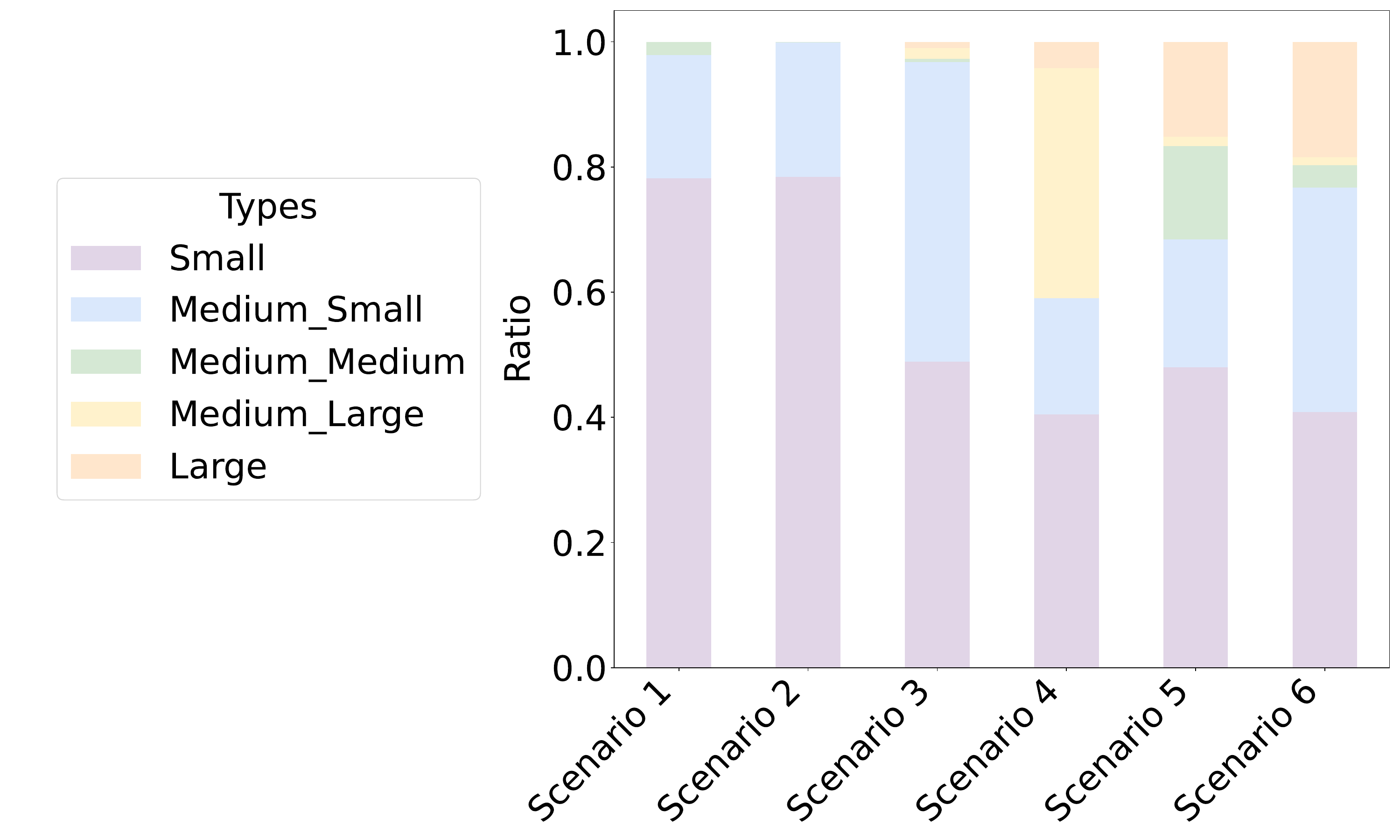}
    \end{subfigure}
    
\FigureNoteStylere{(a) VM Type Classification; (b) Scenario Features in Huawei Dataset. VMs are classified into five categories based on their CPU and memory request sizes: small, medium, and large, along with two additional groups for cases where CPU and memory demands are misaligned, as illustrated in Figure~\ref{fig:vmtype-scenario}\subrefparen{fig:4.1sub1}. The distinct distribution features of the six scenarios are depicted in \ref{fig:vmtype-scenario}\subrefparen{fig:4.1sub2}.}
\end{figure}

\paragraph{Data Description.} The \textit{Huawei-East-1} dataset\endnote{The GitHub URL for the \textit{Huawei-East-1} partially dataset is \url{https://github.com/huaweicloud/VM-placement-dataset}. } spans over a year, containing approximately 125,000 VM requests, each characterized by key attributes including VM ID, CPU and memory demands, arrival time, and request type(created or released).
Reviewing Figure \ref{fig: sched illu}, which highlights the nonstationary characteristics of VM requests, we divide the VM sequences into six equal-length scenarios in chronological order. Each scenario exhibits distinct VM distribution characteristics, as detailed in Figure~\ref{fig:vmtype-scenario}. For instance, Scenario 1 shows a dominant proportion of Small VMs, while Scenario 5 has a notable spike in Medium\_Large VMs.  After dividing the full dataset into six scenarios, we further partition each scenario into six evenly spaced starting points, and sample VM sequences starting from the first five starting points as the training set, while sequences from the sixth starting point are used as the test set. This yields a deterministic 5:1 split that is chronological rather than random. Additionally, we validate our algorithm in the \textit{AzurePublicDatasetV2} dataset\endnote{The GitHub URL for the \textit{AzurePublicDatasetV2} dataset is \url{https://github.com/Azure/AzurePublicDataset/blob/master/AzurePublicDatasetV2.md}.}, and the results are presented in Appendix \ref{secA2}.

\paragraph{Baselines.} Our baseline methods include both traditional heuristics (i.e., Best-Fit, First-Fit, Hindsight) and learning-based approaches (i.e., SchedRL). Best-Fit selects the server with the highest current allocation rate to maximize resource utilization \citep{johnson1973near}. First-Fit assigns VMs to the first available server based on index order, without considering load distribution \citep{johnson1973near}. Hindsight sorts VM requests by descending lifetime and allocates each to the most suitable machine, minimizing idle periods \citep{sinclair2023hindsight}. SchedRL leverages reinforcement learning for multi-NUMA VM scheduling, formulating the problem as a structured combinatorial optimization task within a learning framework \citep{sheng2022learning}. Additionally, the exact solution obtained by solving the mixed-integer programming using Gurobi will serve as the offline optimal solution \citep{gurobi}.

\paragraph{Environmental Parameters and Algorithm Parameters.} To facilitate result reproduction, we provide environmental parameters (i.e., settings) and algorithm parameters in Table \ref{tab:parameter_list}. The settings are consistent with the preceding motivation experiments in Section \ref{sec: 3.1}.

\begin{table}[htbp]
\centering
\caption{Environmental Parameters and Algorithm Parameters List}\label{tab:parameter_list}
\normalsize
\begin{tabular}{lc}
    \toprule
    \textbf{Parameter} & \textbf{Value} \\
    \midrule
    \multicolumn{2}{l}{\textbf{Environmental Parameters}} \\
    \midrule
    Number of PMs \(N\) & 50 \\
    Language agent \(\mathcal{LLM}\) & GPT-4 \\
    \midrule
    \multicolumn{2}{l}{\textbf{General Algorithm Parameters}} \\
    \midrule
    Temperature \(\xi\) & 0.8 \\
    Top-$M$ & 2 \\
    Transfer step of option \(\tau_{\text{max}}\) & 50 \\
    Number of initial options \(K\) (corresponding to scenarios) & 6 \\
    Number of independent experiments $n_s$ & 30 \\
    \midrule
    \multicolumn{2}{l}{\textbf{Option Miner Parameters}} \\
    \midrule
    Iterations in \textit{Option Miner} \(M_{M_i}\) & 300 \\
    Seed heuristic \(\pi^{(0)}\) & Best-fit algorithm \\
    Tokens of each iteration in \textit{Option Miner} & 1,000 \\
    \midrule
    \multicolumn{2}{l}{\textbf{Option Composer Parameters}} \\
    \midrule
    Iterations in \textit{Option Composer} \(M_{C_o}\) & 300 \\
    Sample length of VM sequence \(L\) (context) & 200 \\
    Tokens of each iteration in \textit{Option Composer} & 1,000 \\
    \bottomrule
\end{tabular}
\end{table}

\paragraph{Performance Metric.} For simplicity, configure the cumulative reward $\mathcal{J}$ in Eq. \eqref{eq: mdp-obj}. 
And the environment terminates after the first unsuccessful allocation. Under this configuration, the cumulative reward directly corresponds to the \textit{scheduled length}, defined as the total number of successfully allocated VMs before termination.

To explore the algorithm limits, we use the \textit{performance ratio} as our metric on the training set. The \textit{performance ratio} is calculated by comparing the scheduled length of the online algorithm to that of the offline algorithm and expressed as a percentage:
\begin{equation}
\label{eq: per-ratio}
    \text{Performance Ratio} = \left( \frac{\text{Scheduled length}_{\text{online}}}{\text{Scheduled length}_{\text{offline}}} \right) \times 100\%.
\end{equation}

Furthermore, to assess the robustness of code generation, we report the \textit{code valid ratio} as an additional evaluation metric. A generated policy implementation is regarded as \textit{valid} if it can be executed without errors and produces correct scheduling results during evaluation. The ratio is computed as:
\begin{equation}
\label{eq: valid-ratio}
\text{Code Valid Ratio} = \left( \frac{\text{Number of valid code samples}}{\text{Total number of generated code samples}} \right) \times 100\%.
\end{equation}

We evaluate the proposed framework in a simulated cloud environment consisting of $N=50$ physical machines, using real VM request traces as the workload dataset. The request sequence is segmented into $K=6$ non-overlapping scenarios, each associated with an initial option policy. In the \textit{Option Miner}, GPT-4 is employed to iteratively generate candidate heuristics for 300 iterations, each consuming 1,000 tokens, while retaining the top-$M=2$ high-performing policies at each iteration. In the \textit{Option Composer}, the retained policies are further recombined across scenarios for another 300 iterations under the same token budget, also keeping the top-$M=2$. The maximum transfer step of an option is set to $\tau_\text{max}=50$, and the VM request context length is fixed at $L=200$. The decoding temperature of the LLM is set to $\xi=0.8$. All experiments are repeated over $n_s=30$ independent runs, with results reported as the mean performance.  This corresponds to an end-to-end offline training time of \(\approx 12\) hours and an API usage cost of \(\approx\) USD \(\$63\).  All experiments are conducted using Python 3.7, implemented on a server configured with Ubuntu 20.04.4 LTS. The computational platform comprised four AMD EPYC 7R32 48-Core processors, totaling 192 cores, 514 GB of RAM, and an NVIDIA A100 GPU.

\begin{table}[htbp]
\centering
\caption{Performance Ratio Comparison of Best-Fit, First-Fit, Hindsight, and MiCo Algorithms Across Different Scenarios.}
\label{tab:performance_all_alg}
{\small
\setlength{\tabcolsep}{4pt}
\resizebox{\textwidth}{!}{%
\begin{tabular}{ccccccccc}

\toprule
  \diagbox{\textbf{Algorithm}}{\textbf{Scenario}}  &  $\mathbf{S}^{\mathbf{1}} $ &  $\mathbf{S}^{\mathbf{2}} $ &  $\mathbf{S}^{\mathbf{3}} $ & $ \mathbf{S}^{\mathbf{4}}  $&  $\mathbf{S}^{\mathbf{5}}  $ &  $ \mathbf{S}^{\mathbf{6}}  $& \textbf{Mean} \\
\midrule
Best-Fit & 100.0\% & 99.3\% & 87.4\% & \textbf{93.2\%} & 74.4\% & 84.9\% & 92.6\% \\
First-Fit & 100.0\% & 99.2\% & 87.2\% & 91.8\% & 63.4\% & 76.5\% & 89.7\%  \\
Hindsight & 99.9\% & 99.2\% & 87.3\% &91.7\% & 67.5\% & 78.0\% & 90.5\%  \\
SchedRL & 99.9(±0.0)\% & 97.8(±0.0)\% & 85.2(±1.8)\% & 77.3(±0.1)\% & 51.0(±6.1)\% & 69.5(±3.0)\% & 85.8(±0.8)\% \\
MiCo & 99.9\% & \textbf{99.4\%} & \textbf{95.3\%} & 92.1\% & \textbf{83.6\%} & \textbf{99.3\%} & \textbf{96.9\%} \\ 
\bottomrule
\end{tabular}%
}}
{\textit{Note.} Due to the randomness inherent in reinforcement learning, the results of SchedRL exhibit randomness, with plus and minus signs indicating the confidence intervals. The mean result is calculated by dividing the average score of the online algorithm across all scenarios by the average score of the offline algorithm across all scenarios, which is effective for evaluating the overall effectiveness of the algorithm. The mean results of the remaining tables are calculated in the same way.}
\end{table}

\subsection{Numerical Results}\label{sec: results}
We present three aspects of evaluation: (1) primary performance outcomes (Section \ref{sec: main results}), ablation experiments (Section \ref{sec: ablation study}), and robustness analysis (Section \ref{sec: robustness analysis}) of MiCo. We adapt the Performance Ratio defined in Eq. \eqref{eq: per-ratio} and the Code Valid Ratio defined in Eq. \eqref{eq: valid-ratio} as the principal performance metrics. To demonstrate the performance gap between our algorithm and the offline optimal solution, we exclusively compute the theoretical upper bounds using the Gurobi optimizer on the training set. The best heuristics generated by MiCo are collected in Appendix \ref{secA3}.

\subsubsection{Performance Analysis of MiCo} \label{sec: main results}

As shown in Table~\ref{tab:performance_all_alg}, our algorithm demonstrates consistent superiority in the dataset. First, our proposed MiCo algorithm surpasses other baseline methods in both overall performance and proximity to the offline upper bound across almost all scenarios, achieving the highest performance ratio (96.9\%) relative to the solution solved by Gurobi. Second, when compared with reinforcement learning approaches, our method exhibits amplified advantages. Specifically, it outperforms SchedRL by 11.1\% in mean performance, with particularly notable gaps in complex scenarios (32.6\% improvement in Scenario 4). 
Third, scenario analysis reveals critical insights. 
In Scenario 1, homogeneous small requests dominate, allowing all algorithms to approach the upper bound (within a 0.1\% gap), as simplified scheduling eliminates the need for strategic differentiation.
In Scenario 5, the presence of heterogeneous workloads (5 VM types) exposes the limitations of baseline algorithms, resulting in a performance degradation of more than 20\%, while MiCo demonstrates high stability.

\begin{figure}[htbp]
  \caption{Scheduled Length Comparison of Best-Fit, First-Fit, Hindsight, SchedRL, and MiCo Algorithms on the Test Dataset.}
  \centering
  \includegraphics[width=0.45\textwidth]{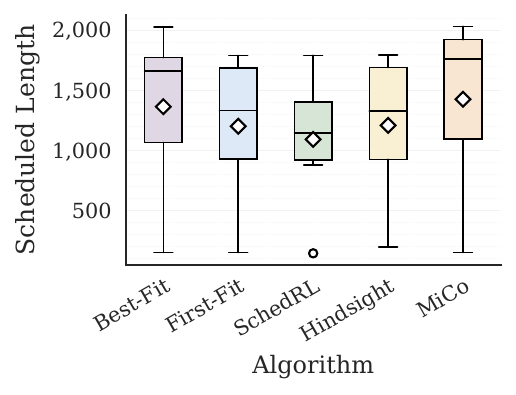}
    \label{fig:test_performance}
  
\end{figure}

To demonstrate the generalization capability of our algorithmic framework, we also sample sequences from the dataset that do not overlap with the training set. These unseen sequences were evaluated, as illustrated in Figure \ref{fig:test_performance}. First, both Best-Fit and MiCo deliver the highest performance; however, MiCo distinguishes itself with the highest median value and box position, signifying consistent outperformance across test scenarios. Second, while First-Fit and Hindsight demonstrate comparable distribution patterns, Hindsight shows improved stability evidenced by its shorter lower whisker. Third, SchedRL exhibits concentrated data distribution around lower scheduled length values, indicating suboptimal overall performance. Collectively, these results validate that MiCo maintains performance superiority even on unseen datasets. 

We also evaluate our algorithm on the Azure public dataset, as shown in Appendix Table \ref{tab:performanceofazure} and Figure \ref{fig:test_performance_azure}. 
The results confirm that our algorithm consistently delivers significant performance improvements across both datasets. Although the Azure dataset has limited variability, our algorithm still outperforms the others.

\subsubsection{Ablation Study} \label{sec: ablation study}
To understand the contribution of each component in MiCo, we conduct a set of ablation studies. First, we examine a baseline configuration in which no scenario generation or contextual characterization is applied, i.e., all instances are optimized in a single undifferentiated search space. Second, we remove the hierarchical policy-switching mechanism and evaluate performance when all decisions must be made by a single unified policy. Finally, we exclude context-aware representations while keeping scenario segmentation, isolating the value of contextual signals within each scenario.

\paragraph{Ablation Study on Scenario Generation.} As shown in Table~\ref{tab:single-heu}, the motivating experiments yield three key insights. 
First, \textit{MiCo w/o Scenario \& Context} achieves only marginal improvement (1.4\%) over the Best-Fit baseline, reflecting the inherent limitation of static heuristics under nonstationary VM request streams. 
Second, \textit{MiCo w/o Scenario}, which augments the policy with contextual inputs while maintaining a single global policy, performs slightly worse (by 0.7\%) than its context-independent counterpart. This indicates that a single context-aware policy may overfit short-term variations without adequately capturing longer-term structural shifts. 
Third, \textit{MiCo w/o Scenario (Per-Trace)} attains the highest performance (96.2\%) by tailoring policies to individual traces. However, this gain comes at the cost of generalization and scalability, rendering it unsuitable for operational settings with evolving and unseen workloads.

\begin{table}[htbp]
\centering
\caption{Performance Comparison of MiCo Variants without Scenario Decomposition.}
\normalsize
\label{tab:single-heu}
\begin{tabular}{l c c}
\toprule
\textbf{Algorithm} & \textbf{Performance Ratio} & \textbf{Gap vs. Best-Fit} \\
\midrule
Best-Fit & 92.6\% & -- \\
MiCo w/o Scenario \& Context & 94.0\% & +1.4\% \\
MiCo w/o Scenario & 93.3\% & +0.7\% \\
MiCo w/o Scenario (Per-Trace) & \textbf{96.2\%} & +3.6\% \\
\bottomrule
\end{tabular}
\end{table}

\begin{table}[!htbp]
\centering
\small
\caption{Performance Ratio and Code Valid Ratio Comparison of Best-Fit, Scenario-Specified Policies ($\text{Policy}^i, i \in \{1,\ldots,6\}$), Random Composer, MiCo without Pruning and MiCo.}

\begin{tabular}{ccccccccc}
\toprule
  \diagbox{\textbf{Algorithm}}{\textbf{Scenario}}  &  $\mathbf{S}^{\mathbf{1}} $ &  $\mathbf{S}^{\mathbf{2}} $ &  $\mathbf{S}^{\mathbf{3}} $ & $ \mathbf{S}^{\mathbf{4}}  $&  $\mathbf{S}^{\mathbf{5}}  $ &  $ \mathbf{S}^{\mathbf{6}}  $& \textbf{Mean} &  \textbf{Code Valid Ratio} \\
\midrule
 Best-Fit & 100.0\% & 99.3\% & 87.4\% & 93.2\% & 74.4\% & 84.9\% & 92.6\% & - \\
 $\textrm{Policy}^{\textrm{1}} $ & \textbf{100.0\%} & 99.3\% & 87.4\% & 93.3\% & 69.7\% & 80.5\% & 91.3\% & 94.7\% \\
 $\textrm{Policy}^{\textrm{2}} $  & 99.9\% & \textbf{99.5\%} & 89.3\% & 92.0\% & 70.4\% & 82.6\% & 91.9\% & 84.0\% \\
 $\textrm{Policy}^{\textrm{3}} $  & 91.3\% & 97.7\% & \textbf{95.5\%} & 93.6\% & 74.5\% & 87.1\% & 90.5\% & 86.4\% \\
 $\textrm{Policy}^{\textrm{4}} $  & 85.6\% & 87.0\% & 84.1\% & \textbf{100.0\%} & 69.7\% & 87.0\% & 84.7\% & 96.6\% \\
$\textrm{Policy}^{\textrm{5}} $  & 90.8\% & 95.4\% & 86.1\% & 91.6\% & \textbf{89.4\%} & 87.0\% & 90.8\% & 79.3\% \\
$\textrm{Policy}^{\textrm{6}} $  & 84.8\% & 86.8\% & 81.7\% & 93.7\% & 79.7\% & \textbf{99.8\%} & 87.4\% & 81.6\% \\
Random Composer & 99.8\% & 97.8\% & 95.0\% & 93.2\% & 74.3\% & 87.0\% & 93.2\%  & - \\
 MiCo w/o-pruning & 99.9\% & 97.7\% & 91.1\% & 99.8\% & 74.6\% & 99.6\% & 95.4\% & 88.4\% \\
 MiCo & 99.9\% & 99.4\% & 95.3\% & 92.1\% & 83.6\% & 99.3\% & \textbf{96.9\%} & 88.4\% \\
\bottomrule
\end{tabular}
\label{tab:impact_of_filter}
\end{table}

\paragraph{Ablation Study on Option Composer.}
To quantify the contribution of the Option Composer, we evaluate the untrained Composer, i.e., Scenario‑Specified Policies, where the Composer defaults to a fixed selection.
As shown in Table \ref{tab:impact_of_filter}, these policies perform best when the scene identifier is known but lack robustness across scenarios. For instance, Policy$^4$ achieves perfect accuracy (100.0\%) in S$^4$ but exhibits a performance drop in cross-scenario evaluations, underscoring the limitations of a fixed Composer. This validates our hypothesis that static policies lack adaptability to dynamic environments, necessitating an intelligent policy switching mechanism like our \textit{Option Composer}. 

Additionally, we further report a Random Composer variant that uniformly switches among the scenario-specified policies. Its performance is lower than that of the policy chosen intelligently (i.e., MiCo) across all scenarios (mean 93.4\% vs 96.9\%). 
These ablation studies quantify the gap before and after Composer training and demonstrate the benefit of context‑aware composition.

\paragraph{Ablation Study on Option Pruning.}
To assess the impact of option pruning, we train the MiCo algorithm without pruning (i.e., MiCo w/o-pruning) for each scenario. After pruning (Eq. \eqref{eq: pruning}), the policy set of MiCo retains the original Policy$^1$, Policy$^2$, Policy$^3$, Policy$^5$, and Policy$^6$. First, as shown in Table \ref{tab:impact_of_filter}, the mean results demonstrate that the performance of the MiCo algorithm is superior to that of the baseline Best-Fit and scenario-specified policies, regardless of whether pruning is applied or not. Second, by using pruning to identify policies that can address multiple scenarios and reduce the selection space, we observe more substantial performance gains (mean 95.4\% vs. 96.9\%). 
However, in S$^4$, pruning leads to a noticeable performance decline (99.8\% vs. 92.1\%). This is because Policy$^4$, which is highly specialized for S$^4$, did not meet the cross-scenario robustness criterion and is therefore excluded. As a result, MiCo has to rely on policies transferred from other scenarios, which only partially overlapped with the structural characteristics of S$^4$.
This illustrates an inherent trade-off: pruning improves efficiency by reducing the policy set, but it may sacrifice scenario-specific optimality, especially in diverse and nonstationary environments.
Such a trade-off also explains why MiCo does not achieve the Best-Fit performance for S$^4$ in Table~\ref{tab:performance_all_alg}. Besides, we further investigate the convergence rates of the proposed algorithm MiCo, the details are shown in Appendix \ref{secA1}.

\subsubsection{Robustness Analysis} \label{sec: robustness analysis}
To examine whether our algorithm is sufficiently robust under nonstationary conditions, we conduct experiments across varying sample lengths, temperatures, and different language agents. The sample length affects the ability of the language model to capture contextual information. The temperature parameter, which controls the stochasticity and creativity of language agents, is varied to assess its impact on the performance of the algorithm. Additionally, we test the performance using multiple LLMs to evaluate their generalizability and robustness across different model architectures.

\begin{figure}[htbp]
  \centering
  \caption{Context Sample Length and Temperature Effects on the Robustness of Algorithm Performance.}
  \vspace{2mm}
  \begin{subfigure}[b]{0.45\textwidth}
    \caption{ }
    \includegraphics[width=\textwidth]{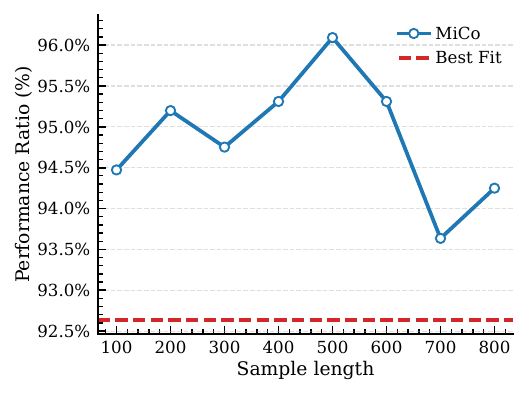}
    \label{fig:sample-length}
  \end{subfigure}
  \hspace{0.05\textwidth}
  \begin{subfigure}[b]{0.45\textwidth}
   \caption{ }
    \includegraphics[width=\textwidth]{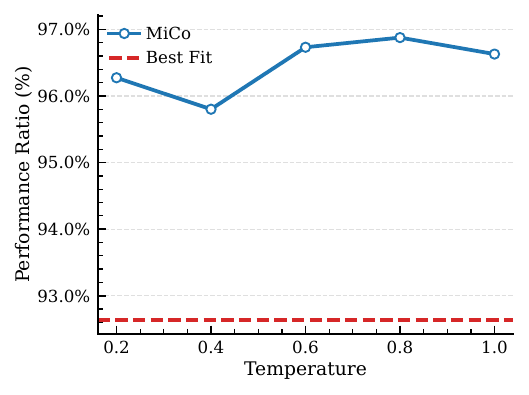}
    \label{fig:temperature}
  \end{subfigure}%
\label{fig:robust-analysis}

\FigureNoteStylere{(a) Impact of Context Sample Length $L$ (Ranging from 100 to 800); and (b) Impact of Temperatures on Algorithm Performance.}
\end{figure}

\paragraph{Robustness Analysis of Context Sample Length.} The context sample length denotes the sequence length of VM traces. We examine sample length $L$ ranging from 100 to 800. The findings from Figure \ref{fig:robust-analysis}\subrefparen{fig:sample-length} indicate that both excessively long (more than 600) and short (less than 400) sample lengths can degrade algorithmic performance, potentially due to the impact of sample length on the ability of the model to extract information. However, our evaluation of different \( L \) values demonstrates that all tested configurations surpass the baseline Best-Fit threshold, indicating the robustness of performance across the sampled range. 

\paragraph{Robustness Analysis of Temperature.}
Temperature is a key factor influencing the stochasticity and creativity of LLMs. We test temperatures $\xi$ of 0.2, 0.4, 0.6, 0.8, and 1.0, as shown in Figure \ref{fig:robust-analysis}\subrefparen{fig:temperature}. While higher temperatures initially enhance output diversity, performance peaks at moderate temperatures (around 0.6 to 0.8), after which it declines. Notably, our framework demonstrates robustness within this range.

\paragraph{Robustness Analysis of LLMs.}
We utilize several LLMs, including Deepseek-Coder-V1, Deepseek-Coder-V2, GPT-3.5-Turbo, and GPT-4, all of which contributed to performance improvements.
As presented in Table \ref{tab:llmperformance}, GPT-4 demonstrates superior overall performance compared to DeepSeek-Coder-V1, DeepSeek-Coder-V2, and GPT-3.5-Turbo.  Notably, GPT-4, GPT-3.5-Turbo, and DeepSeek-Coder-V2 achieve substantially higher code accuracy rates relative to DeepSeek-Coder-V1. This trend suggests that larger models, equipped with a greater number of parameters, tend to yield improved code accuracy.

\begin{table}[htbp]
\centering
\hspace{-2cm}
\footnotesize
\caption{Performance Ratio and Code Valid Ratio across Different LLMs, including Deepseek-Coder-V1, Deepseek-Coder-V2, GPT-3.5-Turbo, and GPT-4.}
\begin{tabular}{ccccccccc}
\toprule
 \diagbox{\textbf{Algorithm}}{\textbf{Scenario}}  &  $\mathbf{S}^{\mathbf{1}} $ &  $\mathbf{S}^{\mathbf{2}} $ &  $\mathbf{S}^{\mathbf{3}} $ & $ \mathbf{S}^{\mathbf{4}}  $&  $\mathbf{S}^{\mathbf{5}}  $ &  $ \mathbf{S}^{\mathbf{6}}  $& \textbf{Mean} &  \textbf{Code Valid Ratio} \\
\midrule

 Best-Fit & 100.0\% & 99.3\% & 87.4\% & 93.2\% & 74.5\% & 84.9\% & 92.6\% & -\\

     MiCo\&DeepSeek-Coder-V1 & 99.9\% & 99.4\% & 92.0\% & 93.3\% & 79.6\% & \textbf{99.4}\% & 96.2\%   & 58.8\% \\

 MiCo\&DeepSeek-Coder-V2 & 99.9\% & \textbf{99.5}\% & 94.9\% & \textbf{99.6}\% & 74.9\% & \textbf{99.4}\% & 96.1\% & 81.7\% \\
 MiCo\&GPT-3.5-Turbo & 100.0\% & 99.3\% & 95.1\% & 93.6\% & \textbf{83.6}\% & 85.1\% & 94.5\% & 70.8\% \\

 MiCo\&GPT-4 & 99.9\% & 99.4\% & \textbf{95.3}\% & 92.1\% & \textbf{83.6}\% & \textbf{99.4}\% & \textbf{96.9}\%  & 88.4\% \\

\bottomrule
\end{tabular}
\label{tab:llmperformance}
\end{table}


\subsection{Interpretable Insights}\label{sec: interpretable insights}
To assess the interpretability of the LLM-generated heuristics, we compare them against established scheduling algorithms. The analysis reveals that several learned policies independently rediscover core design principles consistent with classical human-devised heuristics, as summarized below. The complete code can be seen in Online GitHub.\endnote{The code for this study is publicly available at \url{https://github.com/Eutopiax/llmForScheduling}.}

\subsubsection{Option Miner Analysis.}

During the Option Miner stage, the LLM produces heuristic rules with different origins (Figure~\ref{fig: Miner Code Snippet}). Some rules directly align with established heuristics in the literature (Existing), some integrate multiple known principles into a unified rule (Combined), while others have no clear counterparts in prior studies and may reflect transferred optimization patterns (Innovative).

\begin{figure}[!htbp]
    \centering
    \caption{Code Snippet: Heuristic Code Generated by Option Miner.}
    \label{fig: Miner Code Snippet}
    \begin{tcolorbox}[colback=gray!5!white, colframe=gray!50!black, title=Existing Heuristic Rules : Tight-fit bonuses / residual minimization, top=0pt, bottom=0pt]
    
    \begin{lstlisting}
\# perfect fill
if cpu_remaining==0 and mem_remaining==0:
    total_score += 100
\# banded bonus for small leftover
remaining_percentage = (cpu_remaining + mem_remaining) / (bin[0]+bin[1])
if remaining_percentage < 0.25: 
    total_score += 10
elif remaining_percentage < 0.5: 
    total\_score += 5
    \end{lstlisting}
\end{tcolorbox}

    \begin{tcolorbox}[colback=blue!5!white, colframe=blue!40!black, title=Combined Heuristic Rules : Weighted Sum + Dynamic Weights + Threshold Bands, top=0pt, bottom=0pt]
    
    \begin{lstlisting}
\# Dynamic Weights 
cpu_weight = 0.5 if resource_imbalance < 0.1 else (0.7 if cpu_utilization < mem_utilization else 0.3)
mem_weight = 1 - cpu_weight
\# Weighted Sum 
priority_score = (cpu_weight * cpu_utilization) + (mem_weight * mem_utilization)
if resource_imbalance <= 0.1:
    priority_score += 0.1
\# Threshold Bands 
if cpu_remaining < 0.1 * bin[0] or mem_remaining < 0.1 * bin[1]:
     priority_score -= 0.1
    \end{lstlisting}
\end{tcolorbox}

    \begin{tcolorbox}[colback=green!5!white, colframe=green!40!black, title=Innovative Heuristic Rules : Multi-threshold Relaxation Curve , top=0pt, bottom=0pt]
    
    \begin{lstlisting}
priority_weights = [0.05, 0.1, 0.15, 0.2, 0.25]
cpu_priority_factor = sum(1 for w in priority_weights if cpu_diff < w * bin_cpu)
mem_priority_factor = sum(1 for w in priority_weights if mem_diff < w * bin_mem)
cpu_slack_penalty = 1 - (cpu_priority_factor / len(priority_weights))
mem_slack_penalty = 1 - (mem_priority_factor / len(priority_weights))
cpu_mem_diff_ratio = abs(cpu_diff - mem_diff) / (bin_cpu + bin_mem)
balance_bonus = 1 - cpu_mem_diff_ratio
\# Combined scoring: fit * balance * slack penalties
score = -(bin_fit_score * balance_bonus * (cpu_slack_penalty + mem_slack_penalty))
    \end{lstlisting}
\end{tcolorbox}
    
\end{figure}

\paragraph{Existing Heuristic Rule : Tight-fit bonuses / residual minimization.}
Classic bin-packing heuristics emphasize minimizing unused capacity, known as \textit{residual minimization} or \textit{tight-fit optimization}~\citep{LodiThesis,HeuristicPlacement2D}. 
This rule assigns discrete bonuses for compact placements: a \textit{perfect fill} yields maximum reward, while smaller residual bands (e.g., $<$25\%, $<$50\%) receive scaled incentives. 
Such stepwise scoring achieves efficient resource usage with minimal computational overhead, forming a practical baseline for multi-resource placement.

\paragraph{Combined Heuristic Rule : Weighted Sum + Dynamic Weights + Threshold Bands.}
This rule merges \textit{weighted-sum aggregation} of CPU and memory utilization~\citep{marler2010weighted} with adaptive weights that respond to resource imbalance~\citep{nehra2023efficient}. 
Dynamic adjustment prioritizes underutilized dimensions, promoting balanced saturation. 
Threshold bands introduce soft penalties for near-empty or overfilled bins, smoothing transitions between feasible and suboptimal states. 
Overall, it maintains stability across heterogeneous workloads while capturing both global utilization and local balance.

\paragraph{Innovative Heuristic Rule : Multi-threshold Relaxation Curve.}
This rule offers a graduated fit quality assessment by comparing resource residuals (slack) to scaled thresholds. It tallies thresholds exceeded by normalized slack, generating a priority factor—higher for tight fits, lower for loose ones.
Slack penalties (1 minus normalized factor) deter excessive CPU/memory residuals, minimizing fragmentation and favoring compactness. The balance bonus promotes proportional resource use, preventing imbalances that hinder placements.
The final score multiplicatively fuses base fit compactness, balance, and summed penalties, advancing ``tight-fit" heuristics into a smooth, self-adaptive strategy for dynamic scheduling.

\paragraph{Mapping Heuristic Strategies to Scenarios.}
We observe that the heuristic rules generated by the Option Miner exhibit scenario-specific adaptation. In Scenario~6 (Figure~\ref{fig:vmtype-scenario}), the workload is dominated by \textit{Small} and \textit{Medium\_Small} tasks, indicating a preference for allocating smaller VMs. This behavior results from optimizing the CPU/memory ratio and matching task-to-bin sizes, thereby preventing small tasks from occupying large resource pools. The learned strategy reflects two complementary principles:  reserving large-capacity bins to avoid overfilling and applying penalties to discourage excessive idle space. Together, these mechanisms enable balanced resource utilization and preserve flexibility for future allocations.

\subsubsection{Option Composer Analysis.} 

\begin{figure}[htbp]
    \centering
    \caption{Code Snippet: Heuristic Code Generated by Option Composer.}
    \label{fig: Composer Code Snippet}
    \begin{tcolorbox}[colback=orange!5!white,
                     colframe=orange!50!black,
                     title=Adaptive~Heuristic~Selector~Implementation,
                     top=0pt,bottom=0pt]
\begin{lstlisting}
policy_map = {
    "small":          1,
    "medium_small":   2,
    "medium_medium":  3,
    "medium_large":   4,
    "large":          4,
}
selected_policy = policy_map.get(dominant_request_type, 2)
\end{lstlisting}
    \end{tcolorbox}
\FigureNoteStylere{The numbers 1, 2, 3, and 4 in the figure represent the pruned policies, namely Policy$^2$, Policy$^3$, Policy$^5$, and Policy$^6$.}
\end{figure}

The \textit{heuristic\_selector} function adaptively adjusts heuristic values based on weighted scores, temporal trends, and request-type dynamics to match current workload characteristics. It maps five request types to four heuristics for adaptive response. As shown in Figure~\ref{fig: Composer Code Snippet}, Heuristic~2 is activated when small requests dominate with stable demand (Scenario~2); Heuristic~3 responds to the prevalence of \textit{Medium\_Small} or mixed types (Scenario~3); Heuristic~5 is applied during transitions toward heavier loads (Scenario~5); and Heuristic~6 corresponds to sustained high-capacity demands dominated by \textit{Medium\_Large} or \textit{Large} requests (Scenario~6). This dynamic mapping enables timely and efficient adaptation to workload variations.

\section{Conclusion}\label{sec: conclusion}
The innovation of the MiCo framework lies in its ability to automate and optimize processes that traditionally rely on human expertise. By combining machine learning, data analysis, and simulation techniques, the framework can produce higher-quality, more adaptive scheduling solutions in less time.
Moreover, the framework is designed with scalability and adaptability in mind, enabling extension to future cloud computing scenarios and emerging technological trends.
Future research may explore domain-specific fine-tuning, self-supervised initialization to mitigate cold-start challenges, and multi-objective optimization approaches that integrate energy efficiency with service-level and utilization goals.


\theendnotes
\bibliographystyle{chicago}
\bibliography{arxiv_version}
\clearpage


%

%
%

\appendix
\section{Detailed Literature Review} \label{secA4}
The Dynamic Multidimensional Bin Packing (DMBP) problem is a complex extension of the classical bin packing problem, characterized by the sequential allocation of multidimensional items that arrive and depart dynamically. This problem requires the efficient scheduling of shared resources such as CPU, memory, and bandwidth. Unlike scenarios involving single-unit capacity occupation, DMBP necessitates intricate multidimensional resource sharing, making it a computationally intractable NP-hard combinatorial optimization problem. To address this, researchers employ optimization-based techniques, learning-based methods, or heuristics to derive approximate solutions \citep{coffman1984approximation}.

\textbf{Optimization-based Algorithms for DMBP.} Offline DMBP assumes that all items to be packed and their sequence are known in advance. It can be formulated as a mixed-integer programming problem, which can be solved exactly using branch-and-bound algorithms \citep{martello2000three}. These algorithms are embedded in commercial solvers such as Gurobi \citep{gurobi}. For large-scale problems, decomposition methods have proven effective in solving DMBP \citep{pisinger2007using, puchinger2007models,cote2021combinatorial}.
In the online setting, only the current item information is known, and items must be packed sequentially in an unknown arrival process. When item sizes are stochastic, the problem becomes even more challenging, and most theoretical research focuses on analyzing the multidimensional and dynamic characteristics separately \citep{chen2023cloud}. For a review of approximation methods for online multidimensional bin packing problems, refer to \cite{christensen2017approximation}. In contrast, much of the literature on online one-dimensional dynamic bin packing focuses on applications such as surgery scheduling, where emergency patients arrive randomly and require immediate service with uncertain operation durations \citep{berg2017fast, zhang2020branch}. From an OM perspective, dynamic bin packing can be viewed as a service system, as discussed by \cite{maguluri2012stochastic}, \cite{stolyar2013infinite}, and \cite{stolyar2021service}.
For instance, \cite{stolyar2013infinite} develops an infinite-server system model with homogeneous servers, allowing arriving VMs to be immediately assigned to a server. A series of studies by \cite{stolyar2013infinite, stolyar2021service} reveal that simple randomized policies could achieve asymptotic optimality as the system scale, measured by the VM arrival rate and total number of customers, increases to infinity.


\textbf{Learning-based Algorithms for DMBP.} With the rapid development of machine learning (ML), researchers have explored its application to DMBP.
One approach is to apply ML directly to solve DMBP. For instance, \cite{zhao2021b} combine deep learning and reinforcement learning (RL) techniques to handle the dynamic nature and physical constraints of online packing, improving space utilization and operational efficiency.
Another approach is to integrate RL with heuristics or mathematical models. RL optimizes the decision-making process for bin packing, where heuristic methods provide initial solutions or baselines. The RL algorithms then learn from these baselines and explore better solutions due to their exploratory nature \citep{zhang2022deep}. Additionally, heuristic searches can provide supplementary reward signals to the RL agent, thereby accelerating the training process. \cite{cai2019reinforcement} propose an approach where the RL agent creates an initial feasible solution, which is subsequently optimized further using simulated annealing.
\cite{jiang2021learning} enhance solution quality by integrating RL with constraint programming (CP) by treating orientation and position as decision variables, in which the RL agent dynamically selects their values during a branch-and-bound search, finally improving solution efficiency within the CP framework.

\textbf{Heuristic Algorithms for DMBP.} To efficiently address the online DMBP problem, heuristic algorithms often derive insights from theoretical analysis to accelerate the search process, solving instances with tens of thousands of items within a limited time \citep{coffman1983dynamic, hadary2020protean}. Common algorithms include First-Fit (allocating the next item to the first bin with sufficient capacity) and Best-Fit (allocating the item to the bin with the smallest remaining capacity that still fits) \citep{garey1976resource}. These algorithms provide performance bounds, making them viable as approximation methods. Notably, \cite{azar2013tight} and \cite {azar2019tight} establish tight bounds for online multidimensional and dynamic DMBP, respectively.
When uncertainty or nonstationarity is involved, \cite{bentley1984some} analyze First-Fit and Best-Fit algorithms for online bin packing with stochastic item sizes. \cite{coffman2001bandwidth} study these algorithms in stochastic process settings, focusing on item arrivals governed by stochastic processes.

\textbf{LLMs for Modeling and Code Generation.} The integration of LLMs into combinatorial optimization has advanced two primary methodologies: automated mathematical modeling and heuristic algorithm generation. 
A critical challenge in optimization lies in transforming textual problem descriptions into formal mathematical formulations. Pioneering work in this domain includes the NL4OPT competition \citep{ramamonjison2023nl4opt}, which establishes a benchmark for translating natural language into mathematical programs using pre-trained language models. Building on this foundation, OptiMUS \citep{ahmaditeshnizi2024optimus} introduce a modular framework to formulate and solve mixed-integer linear programming (MILP) problems from natural language inputs, demonstrating end-to-end capabilities from problem interpretation to solver code generation. Further extending the scope, \cite{huang2025orlm} propose a synthetic data generation framework to train LLMs across different optimization problem types, effectively addressing data scarcity challenges in specialized domains.

Beyond mathematical modeling, LLMs exhibit remarkable potential in designing heuristic algorithms for combinatorial optimization problems. For instance, \cite{romera2024mathematical} and \cite{ye2024reevo} leverage LLMs to iteratively generate and refine heuristic rules, outperforming traditional evolutionary algorithms in generalization capability and interpretability for problems like bin packing. Notably, these methods exploit the code generation proficiency of LLMs to evolve context-aware strategies through systematic prompt engineering and fitness-guided evolutionary iterations.

\begin{figure}[htbp]
    \centering
    \caption{Method Comparison of Heuristic, Reinforcement Learning, and Language Agents Methods.}
    \includegraphics[width=1\linewidth]{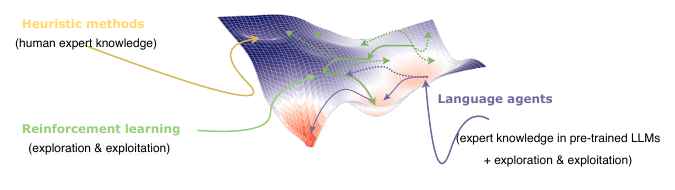}
    \label{fig:method-comparison}
    \par\vspace{-1cm}
\end{figure}

\textbf{Limitations of Existing Methods and the Role of LLMs.} Optimization-based approaches provide theoretical guarantees but face significant computational scalability challenges when addressing large-scale problems with multidimensional and dynamic characteristics. Learning-based methods, particularly RL, offer adaptive decision-making capabilities but typically require substantial computational resources for training. Traditional heuristic methods, while computationally efficient, are susceptible to local optima and heavily rely on manually designed rules based on domain expert knowledge. To address these limitations, we propose integrating the generalized expert knowledge embedded in LLMs with the exploration-exploitation mechanisms of RL frameworks. As illustrated in Figure \ref{fig:method-comparison}, this approach aims to automate the design of hyper-heuristic algorithms, reducing reliance on human experts while adapting to dynamic environments. Our work applies this paradigm to Virtual Machine (VM) scheduling, transitioning from synthetic benchmarks to real-world industrial scenarios with dynamic resource demands and heterogeneous hardware constraints.

\section{Supplementary Experiments}\label{secA2}

\begin{figure}[htbp]
    \centering
    \caption{Characteristics of VM requests in Azure Dataset.}
    \label{fig:VM-distribution-azure}
    \vspace{2mm}
    \begin{subfigure}[b]{0.32\textwidth}
        \caption{\footnotesize  The Number of VM Arrivals. }
        \label{fig:arrival-sub1}
        \includegraphics[width=\textwidth]{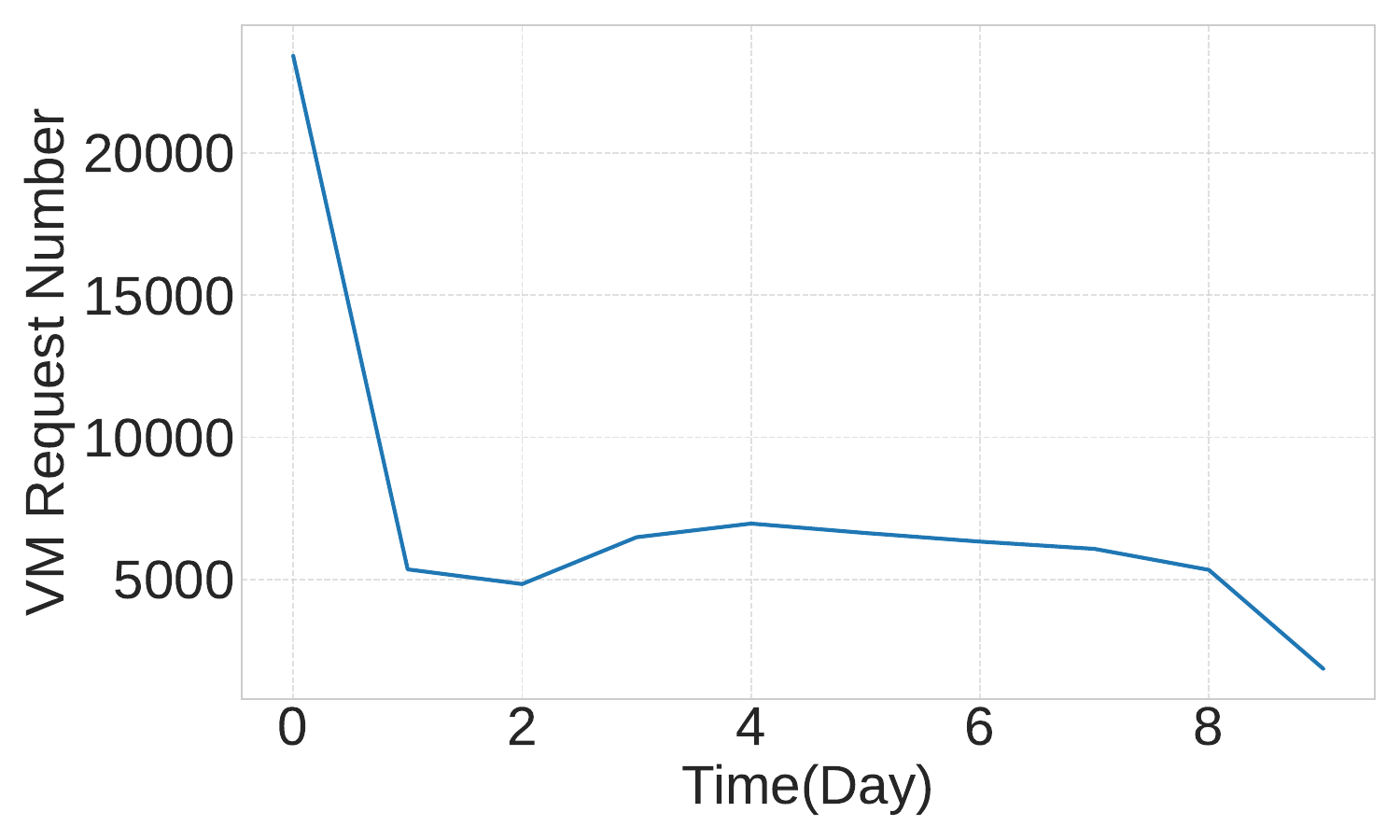}
    \end{subfigure}
    \begin{subfigure}[b]{0.32\textwidth}
        \caption{\footnotesize  Average Duration Time. }
        \label{fig:duration-sub2}
        \includegraphics[width=\textwidth]{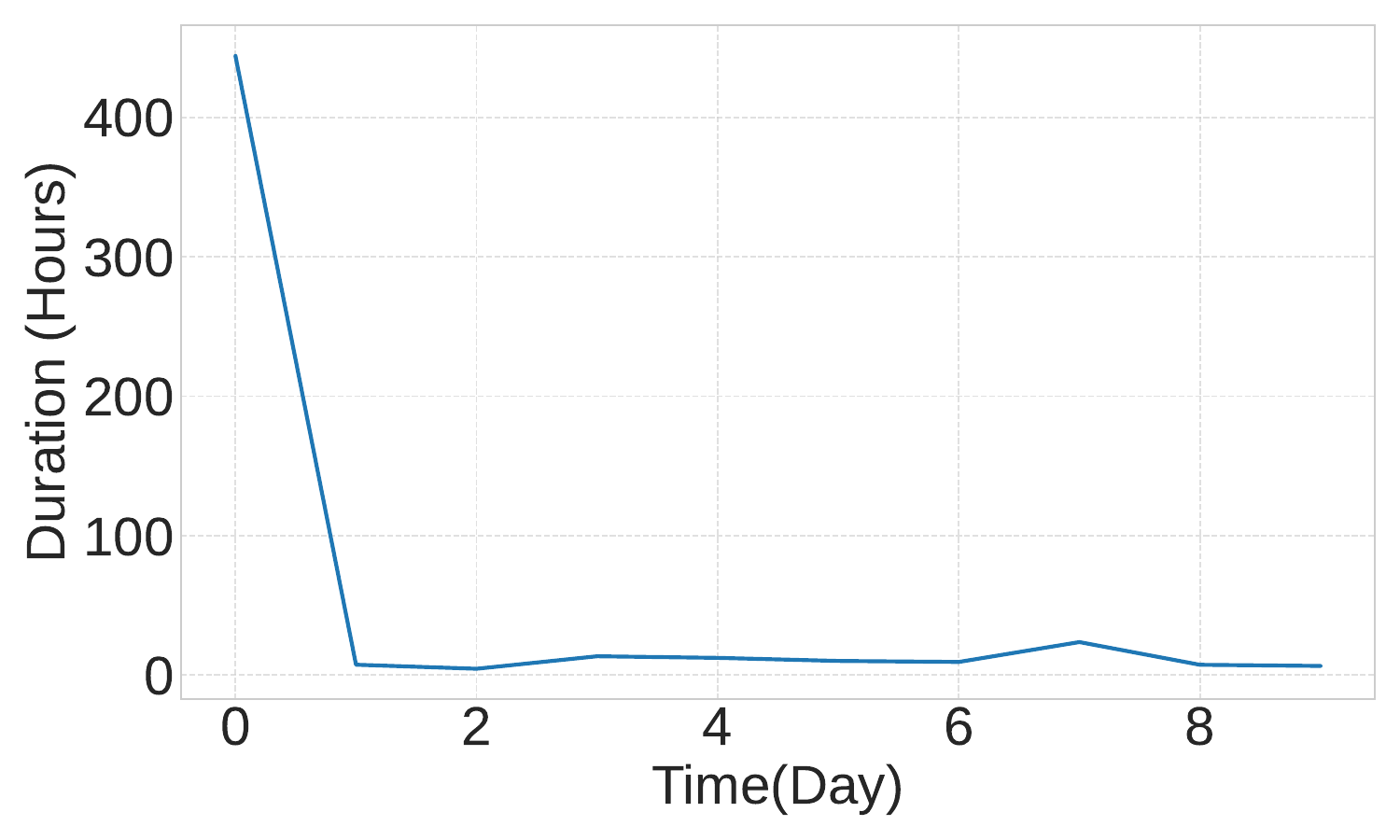}
    \end{subfigure}
    \begin{subfigure}[b]{0.32\textwidth}
        \caption{\footnotesize  Diverse Demand Size Distributions.}
        \label{fig:size-sub3}
        \includegraphics[width=\textwidth]{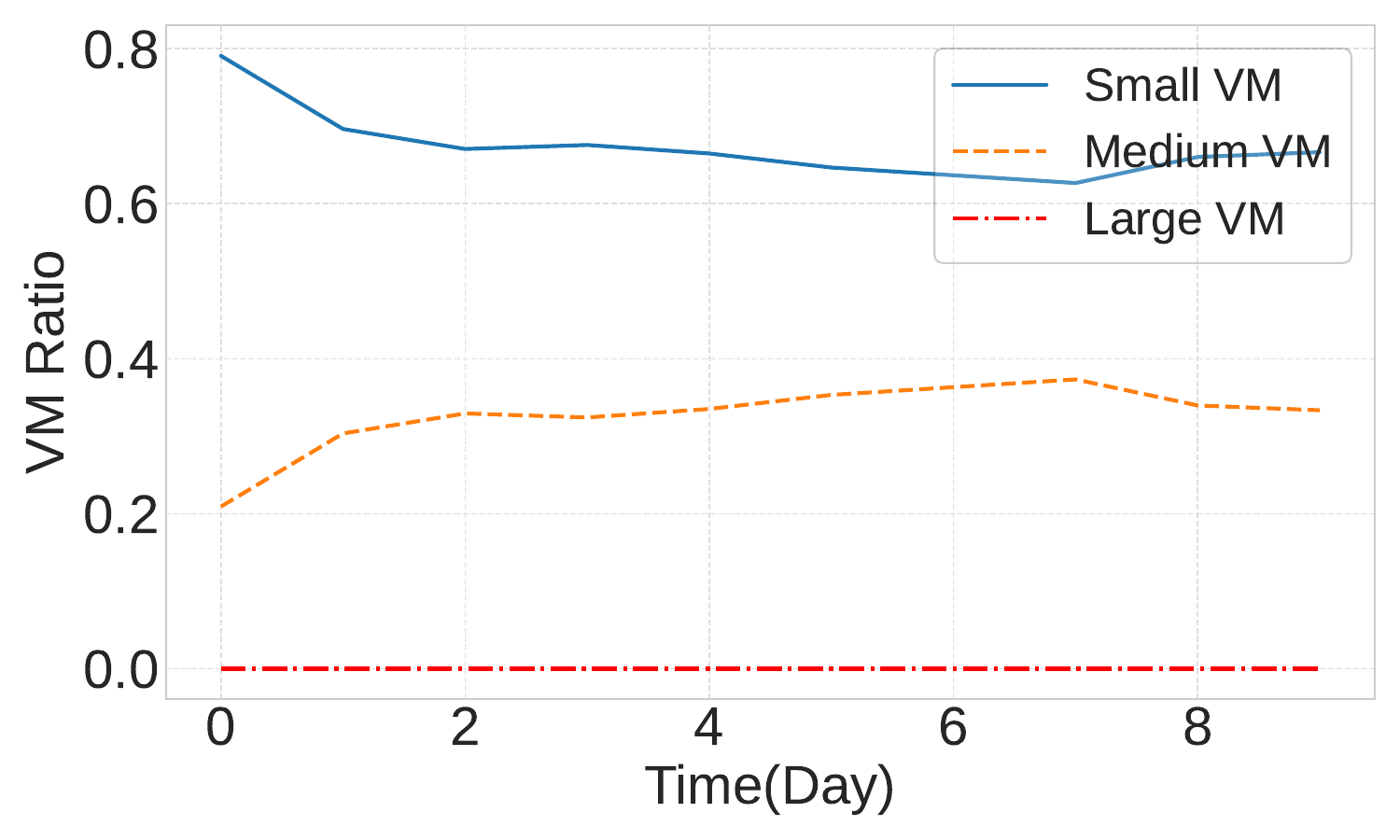}
    \end{subfigure}
    
\end{figure}

\begin{figure}[htbp]
\caption{Scenario Features and Test-set Performance for Each Algorithm on the Azure Dataset.}
\vspace{-5mm}
  \begin{subfigure}[t]{0.54\textwidth}
    \centering
    \caption{\small Scenario Features on the Azure Dataset.}
    \includegraphics[width=\linewidth]{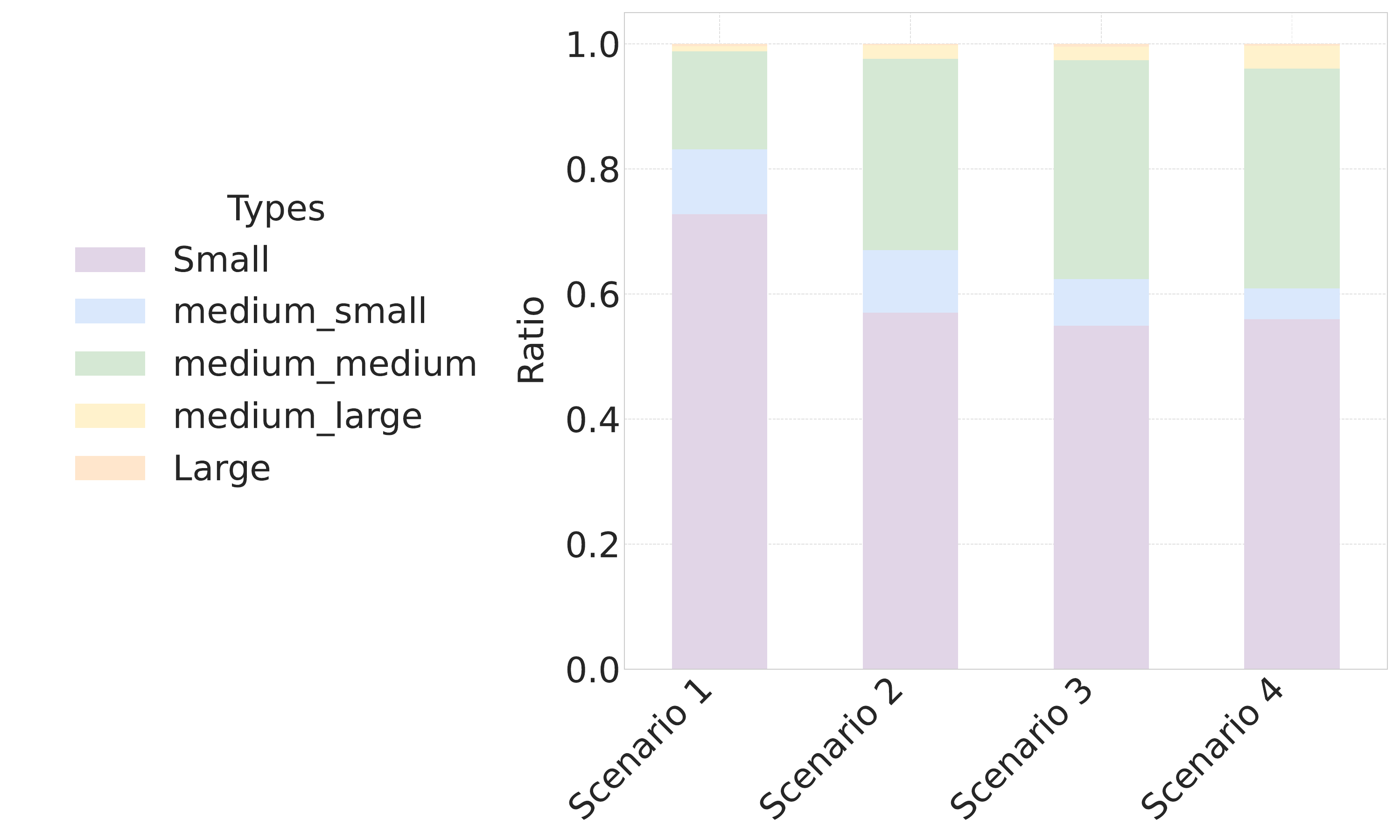}
    \label{fig:scenario_azure}
  \end{subfigure}
  \hfill
  \begin{subfigure}[t]{0.42\textwidth}
    \centering
    \caption{\small Scheduling Length Comparison on the Azure Test dataset.}
    \includegraphics[width=\linewidth]{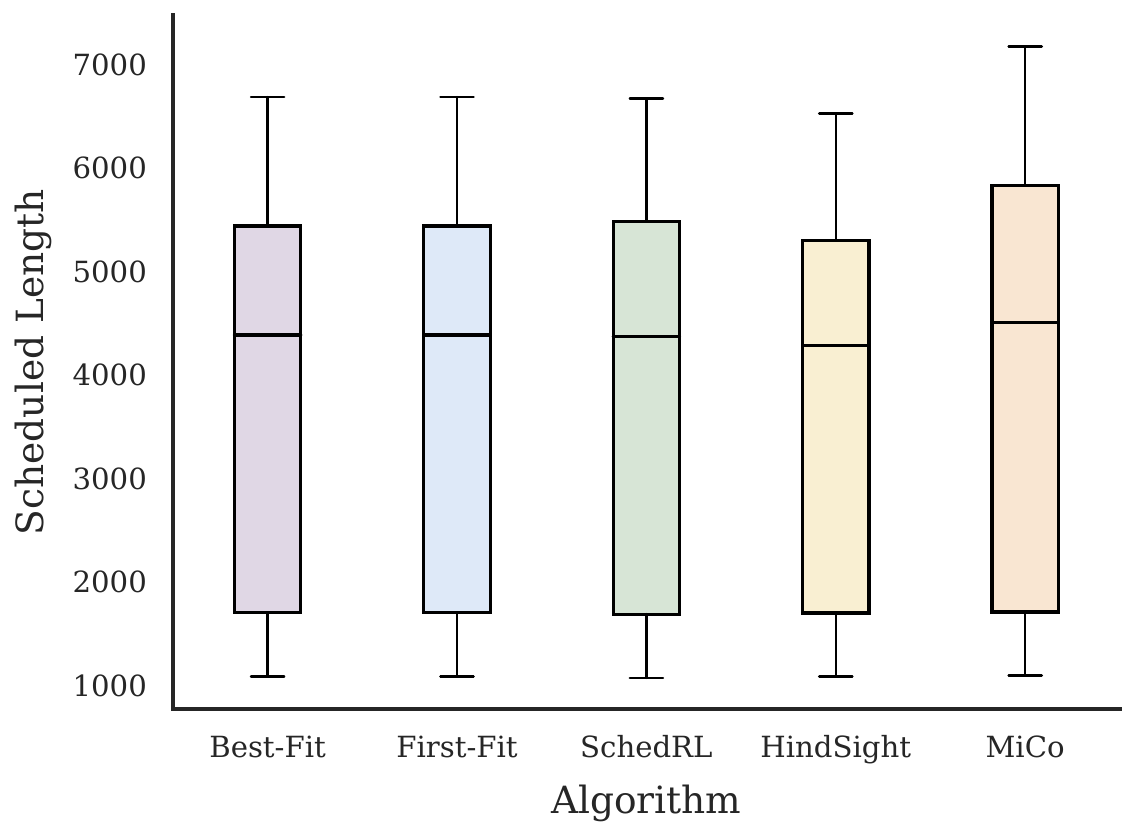}
    
    \label{fig:test_performance_azure}
  \end{subfigure}

  \label{fig:azure_combined}
\end{figure}

We conduct synchronous experimental validation on the Azure dataset.
The dataset \textit{AzurePublicDatasetV2} contains nearly 400{,}000 VM requests and covers approximately one month.
To keep the workload scale comparable to \textit{Huawei-East-1}, we use a chronological, contiguous subset: the first 200{,}000 VM requests ordered by time, which correspond to roughly a 9-day period.
We retain only CPU and memory attributes for Azure to enable a direct comparison of VM characteristics and performance metrics across the two datasets. Notably, we extract some VM sequences from the Azure dataset and find that these sequences lack sufficiently different scene features and are primarily composed of small and medium-sized requests, as shown in Figure \ref{fig:VM-distribution-azure}. As this dataset contains only several days of VM request records and demonstrates no statistically significant nonstationary characteristics in request distribution, we partition it into four scenarios, as shown in Figure \ref{fig:azure_combined}\subrefparen{fig:scenario_azure}. 

We train the MiCo algorithm on the Azure training set and evaluate it on the test set. Experimental results (detailed in Table \ref{tab:performanceofazure}) reveal two principal findings. First, MiCo consistently achieves superior scheduling performance across all scenarios. Particularly in Scenario 2, which features highly diverse VM requests, the algorithm exhibits remarkable effectiveness with a 17.3\% performance gap compared to the heuristic Hindsight solution (70.3\% vs. 53.0\%). This represents an average improvement of about 6\% over existing online scheduling algorithms, conclusively demonstrating the robustness advantages of MiCo in dynamic environments. Second, Scenarios 3 and 4 reveal systemic vulnerabilities when handling medium/large VM requests: Inappropriate resource reservation strategies increase placement failure probabilities, causing premature scheduling termination due to insufficient capacity for subsequent large-task allocations. But this may also be due to the fact that our experimental setup is limited to 50 PMs, which poses inherent challenges in reserving sufficient resources for sequential large-task allocations.

\begin{table}[htbp]
\centering
\small
\caption{Performance Comparison of Best-Fit, First-Fit, Hindsight, and MiCo Algorithms against Offline Solutions Obtained by Gurobi 
Across Scenarios in Azure Training Dataset.\color{red}
}
\begin{tabular}{ccccccc}
\toprule
\diagbox{\textbf{Algorithm}}{\textbf{Scenario}}  &  $\mathbf{S}^{\mathbf{1}} $ &  $\mathbf{S}^{\mathbf{2}} $ &  $\mathbf{S}^{\mathbf{3}} $ & $ \mathbf{S}^{\mathbf{4}}   $& \textbf{Mean} \\
\midrule
 Best-Fit  & 99.2\% & 58.0\% & 57.5\% & 83.1\% & 65.5\% \\
 First-Fit & 99.2\% & 58.2\% & 57.6\% & 83.1\% & 65.6\% \\
 Hindsight & 99.2\% & 53.0\% & 58.2\% & 78.2\% & 62.8\%  \\
 SchedRL  & 98.3(±0.3)\% &   58.1(±0.1)\% &  57.6(±0.2)\% &   83.5(±0.7)\% &  65.7(±0.3)\%   \\
 MiCo$_{\text{transfer}}$ & \textbf{-} & \textbf{-} & \textbf{-} & \textbf{-} & \textbf{63.5\%} \\
  FunSearch & \textbf{-} & \textbf{-} & \textbf{-} & \textbf{-} & \textbf{70.6\%} \\
 MiCo  & \textbf{99.9\%} & \textbf{70.3\%} & \textbf{60.4\%} & \textbf{86.5\%} & \textbf{71.5\%} \\
\bottomrule
\end{tabular}
\label{tab:performanceofazure}
\end{table}

We further include a transfer setting, MiCo trained on Huawei and evaluated on Azure (MiCo$_{\text{transfer}}$), using the same experimental setup as the original Azure experiments, this appears as row 5 in Table \ref{tab:performanceofazure} and performs poorly. The degradation stems from a data mismatch: as shown in Figure \ref{fig:azure_combined}\subrefparen{fig:scenario_azure}, Azure lacks the scenario structure present in Huawei, so a hierarchy mined on Huawei does not transfer well. In addition, we add Context-Independent FunSearch as a baseline on Azure, because Azure is relatively stationary, MiCo trained on Azure yields only marginal improvements over direct FunSearch, consistent with the limited benefits of hierarchical composition in near-stationary settings.
Additionally, we evaluate the algorithm on the test dataset. As shown in Figure \ref{fig:azure_combined}\subrefparen{fig:test_performance_azure}, while the box plot shows similar median performance across all algorithms, the upper whisker of MiCo is longer and its box is positioned relatively higher. This further highlights the superiority of MiCo over the other algorithms.

\begin{figure}[htbp]
\caption{Testset Performance for Each Algorithm on the Huawei Dataset.}
  \vspace{-5mm}
  \begin{subfigure}[t]{0.48\textwidth}
    \centering
    \caption{\small Performance in Huawei Test Dataset (50 servers).}
    \includegraphics[width=\linewidth]{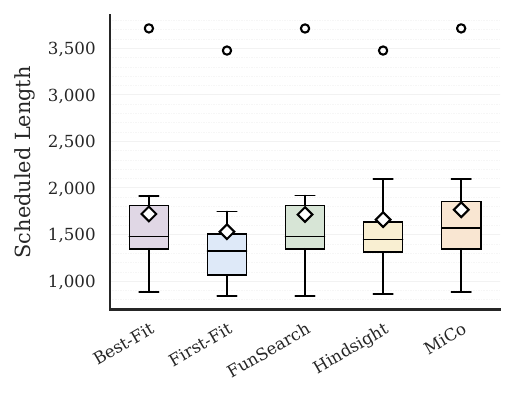}
    \label{fig:test_performance_50}
  \end{subfigure}
  \hfill
  \begin{subfigure}[t]{0.48\textwidth}
    \centering
    \caption{\small Performance in Huawei Test Dataset (100 servers).}
    \includegraphics[width=\linewidth]{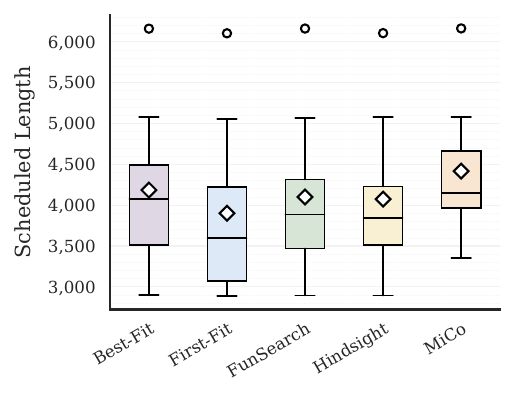}
    
    \label{fig:test_performance_100}
  \end{subfigure}

  \label{fig:test_performance_huawei}
\end{figure}

We conduct an additional evaluation using a data-driven scenario construction and a chronological train–test split. The Huawei dataset is split chronologically, with the first 100,000 samples used for training and the subsequent 25,000 samples for testing. Scenarios are constructed on the training set by first learning a policy for each demand sequence independently, then evaluating these policies across sequences, and clustering them based on performance similarity. This procedure yields five representative scenarios. Figures~\ref{fig:test_performance_huawei}\subrefparen{fig:test_performance_50} and~\ref{fig:test_performance_huawei}\subrefparen{fig:test_performance_100} report test performance under 50-server and 100-server configurations, respectively. The policy trained under the 50-server setting is directly applied to the 100-server system without retraining. In both cases, our method consistently outperforms the baselines, demonstrating robustness to scenario construction and system scale.

\clearpage
\section{Prompt Template}\label{secA1}

\vspace{5mm}
\begin{tcolorbox}[
    title=Scheduler Template,
    label=box: 1, 
    colback=gray!20,      
    colframe=black,       
    width=\textwidth,     
    arc=5mm,              
    boxrule=0.5mm,        
    coltitle=black,       
    toptitle=0mm,         
    colbacktitle=gray!20, 
    fonttitle=\bfseries,
    fontupper=\linespread{1.0}\selectfont 
]

Given the existing priority\_v0 function, please generate an optimized version named priority\_v*. This new version should be more complex and efficient, incorporating multiple conditional logic and loops as necessary. The function should calculate priorities for items to be added to bins, considering the item size and bin capacities. Ensure the function is significantly different and more advanced than the prior versions. Only the Python code for the function is required, without any additional descriptions or annotations. Existing priority\_v0 function for reference:

\begin{lstlisting}[label=box:1, basicstyle= \scriptsize]
def priority_v0(bin, item):
    """ Calculate and return the priority score for adding a specific item to a bin based on available CPU and MEM resources.
    Args:
        bin (tuple): Tuple representing the bin's available resources, where bin[0] is CPU and bin[1] is memory.
        item (tuple): Tuple representing the item's resource requirements, where item[0] is CPU and item[1] is memory needed.
    Returns:
        int: The total score for placing the item in the current bin. A higher score indicates a better fit based on current available resources. """
    score = -(bin[0] - item[0])
    return score
def priority_v1(bin, item):
    """Improved version of `priority_v0`."""
\end{lstlisting}
Your task is to create the optimized priority\_v* function based on the guidelines above. Remember, only the Python function code is needed.

\end{tcolorbox}

\begin{tcolorbox}[
    title=Context-Aware Scheduler
Template,
    label=box: 2, 
    colback=gray!20,      
    colframe=black,       
    width=\textwidth,     
    arc=5mm,              
    boxrule=0.5mm,        
    coltitle=black,       
    toptitle=0mm,         
    colbacktitle=gray!20, 
    fonttitle=\bfseries,
    fontupper=\linespread{1.0}\selectfont 
]

You are a leading expert on this topic. Given the existing heuristic\_selector\_v0 function, please generate an optimized version named heuristic\_selector\_v*. This new version should be more complex and efficient, incorporating multiple conditional logic and loops as necessary. Ensure the function is significantly different and more advanced than the prior versions. Existing heuristic\_selector\_v0 function for reference:

\begin{lstlisting}[label=box:2, basicstyle=\scriptsize]
def heuristic_selector_v0(condition):
    """ This function selects the appropriate heuristic scheduling function based on the input condition.param condition: list of dicts, each representing the distribution of request types over the past 200 requests, divided into 4 groups of 50 requests each.Each dictionary contains keys "small", "medium_small", "medium_medium", "medium_large", and "large"  with their respective proportions.Example:
    [{"small": 0.4, "medium_small": 0.3, "medium_medium": 0.1, "medium_large": 0.1, "large": 0.1},
    {"small": 0.4, "medium_small": 0.3, "medium_medium": 0.1, "medium_large": 0.1, "large": 0.1},
    {"small": 0.4, "medium_small": 0.3, "medium_medium": 0.1, "medium_large": 0.1, "large": 0.1},
    {"small": 0.4, "medium_small": 0.3, "medium_medium": 0.1, "medium_large": 0.1, "large": 0.1}]
    :return: int, index of the selected heuristic function (only 1,2,3,4)"""
    return 1
def heuristic_selector_v1(condition):
    """Improved version of `heuristic_selector_v0`."""
\end{lstlisting}

The ``heuristic\_selector" function should only return 1 or 2 or 3 or 4. In order to ensure the result, do not use any "random" in ``heuristic\_selector" function. Your task is to create the optimized heuristic\_selector\_v* function based on the guidelines above. Remember, only the Python function code is needed.

\end{tcolorbox}

\clearpage
\section{Generated Heuristics}\label{secA3}
\vspace{5mm}

\begin{lstlisting}[language=Python, style=mystyle, caption={Heuristic Algorithm Generated by Option Miner}, label={code:option-miner}]
def priority(bin, item):
    # Unpack CPU and memory resources
    cpu_resource, mem_resource = bin
    cpu_needed, mem_needed = item

    # Return -inf if item doesn't fit
    if cpu_needed > cpu_resource or mem_needed > mem_resource:
        return float('-inf')

    # ... Calculate various differences, ratios, and a weight_factor ...
    cpu_diff = cpu_resource - cpu_needed
    mem_diff = mem_resource - mem_needed
    ...
    weight_factor = 1 + ...

    # ... Loop to calculate the core min_weighted_diff ...
    min_weighted_diff = float('inf')
    for i in range(2):
        # ... Complex weighted_diff calculation ...
        weighted_diff = -((resource_diffs[0] + resource_diffs[1]) * ... * weight_factor)
        
        # ... Conditional adjustment ...
        if ...:
            weighted_diff *= ...

        min_weighted_diff = min(min_weighted_diff, weighted_diff)

    # ... Calculate remaining space and item/bin size ratio ...
    remaining_space = (cpu_diff * mem_diff) / (cpu_resource * mem_resource)
    min_weighted_diff *= (1 - remaining_space)

    ...
    size_ratio = (cpu_needed * mem_needed) / (cpu_resource * mem_resource)

    # ... Select weights based on size_ratio ...
    weighted_items = [0.7, 0.3] if size_ratio <= 0.5 else [0.6, 0.4]

    # ... Define helper function and calculate initial score ...
    def score_by_weight(weight):
        return min_weighted_diff * weight
    ...
    final_score = score_by_weight(best_weight)

    # ... Apply multiple adjustments, penalties, and bonuses to the final_score ...
    final_score += abs(size_ratio - 1) * 0.2
    if ...: # Penalty condition
        final_score *= ...
    if ...: # Bonus condition 1
        final_score *= ...
    if ...: # Bonus condition 2
        final_score *= ...
    final_score *= ... # Final factor adjustment

    # Return the computed final score
    return final_score
\end{lstlisting}

\begin{lstlisting}[language=Python, style=mystyle, caption={Heuristic Algorithm Generated by Option Composer}, label={code:option-composer}]
def heuristic_selector(condition):
    import numpy as np

    # --- Step 1: Calculate statistics ---
    def calculate_stats(condition):
        stats = {
            "averages": {k: 0 for k in condition[0].keys()},
            "trends": {k: [] for k in condition[0].keys()},
            "accelerations": {k: [] for k in condition[0].keys()},
            "recent_changes": {k: 0 for k in condition[0].keys()},
            "recent_accelerations": {k: 0 for k in condition[0].keys()},
        }
        ...
        # compute averages, trends, accelerations
        ...
        return stats

    # --- Step 2: Weighted score calculation ---
    def calculate_weighted_scores(stats):
        weights = [0.25, 0.3, 0.3, 0.15]
        scores = {}
        for k in stats["averages"].keys():
            scores[k] = (
                weights[0] * stats["averages"][k]
                + weights[1] * np.std(stats["trends"][k]) / (np.mean(stats["trends"][k]) + 1e-6)
                + weights[2] * condition[-1][k]
                + weights[3] * np.std(stats["accelerations"][k]) / (np.mean(stats["accelerations"][k]) + 1e-6)
            )
        return scores

    stats = calculate_stats(condition)
    scores = calculate_weighted_scores(stats)

    # --- Step 3: Heuristic selection based on dominant metric ---
    dominant_request_type = max(scores, key=scores.get)
    heuristic_map = {"small": 1, "medium_small": 2, "medium_medium": 3, "medium_large": 4, "large": 4}
    selected_heuristic = heuristic_map.get(dominant_request_type, 2)

    # --- Step 4: Refinement rules ---
    if selected_heuristic == 4 and scores["medium_medium"] > scores["medium_small"]:
        selected_heuristic = 3
    elif selected_heuristic == 1:
        ...
    if stats["trends"]["large"][-1] > (np.mean(stats["trends"]["large"]) + 1.4 * np.std(stats["trends"]["large"])):
        selected_heuristic = 4

    recent_changes = np.array(list(stats["recent_changes"].values()))
    if np.sum(recent_changes ** 2) > 1.2 * len(recent_changes):
        ...

    recent_accelerations = np.array(list(stats["recent_accelerations"].values()))
    if np.sum(recent_accelerations ** 2) > 1.5 * len(recent_accelerations):
        ...

    balanced = [abs(scores[k] - scores["large"]) < 0.1 for k in scores.keys() if k != "large"]
    if all(balanced):
        ...

    if selected_heuristic < 1 or selected_heuristic > 4:
        selected_heuristic = 2

    return selected_heuristic
\end{lstlisting}

\end{document}